\documentclass[runningheads]{llncs}
\usepackage[T1]{fontenc}
\usepackage{graphicx}

\usepackage[outdir=./fig/]{epstopdf}
\usepackage[english]{babel}
\usepackage{amsfonts}
\usepackage{amssymb}
\usepackage{amsmath}
\usepackage{mathtools}
\usepackage[ruled,linesnumbered,noend]{algorithm2e}
\usepackage{booktabs}
\usepackage{nicefrac}
\usepackage{multirow}
\usepackage{enumerate}
\usepackage{tabularx}

\DeclareMathOperator*{\argmax}{argmax}
\DeclareMathOperator{\GNN}{GNN}

\DeclareMathOperator{\MLP}{MLP}

\DeclareMathOperator{\MEAN}{MEAN}
\DeclareMathOperator{\MAX}{MAX}
\DeclareMathOperator{\IR}{\mathbb{R}}
\renewcommand{\phi}{\varphi}

\usepackage{lscape}
\usepackage[pass]{geometry}

\usepackage{hyperref}
\usepackage{color}

\begin{document}
\title{
Learning to Control Local Search for Combinatorial Optimization
}

\author{
Jonas K. Falkner
\and
Daniela Thyssens
\and
Ahmad Bdeir
\and
Lars Schmidt-Thieme
}
\authorrunning{Falkner et al.}

\institute{
Institute for Computer Science, University of Hildesheim, Hildesheim, Germany\\
\email{\{falkner,thyssens,bdeir,schmidt-thieme\}@ismll.uni-hildesheim.de}
}

\maketitle              

\begin{abstract}
Combinatorial optimization problems are encountered in many
practical contexts such as logistics and production, but exact
solutions are particularly difficult to find and usually NP-hard
for considerable problem sizes. To compute approximate solutions,
a zoo of generic as well as problem-specific variants of local
search is commonly used. However, which variant to apply to which particular problem is
difficult to decide even for experts.

In this paper we identify three independent algorithmic aspects
of such local search algorithms and formalize their sequential
selection over an optimization process as Markov Decision
Process (MDP). We design a deep graph neural network as policy model
for this MDP, yielding a learned controller for local search
called NeuroLS. Ample experimental evidence shows that NeuroLS is able to
outperform both, well-known general purpose local search controllers
from Operations Research as well as latest machine learning-based
approaches.

\end{abstract}

\section{Introduction}

Combinatorial optimization problems (COPs) arise in many research areas and applications. They appear in many forms and variants like vehicle routing\cite{toth2002vehicle}, scheduling\cite{taillard1993benchmarks} and constraint satisfaction\cite{tsang2014foundations} problems but share some general properties. One of these properties is that most COPs are proven to be NP-hard which makes their solution very complex and time consuming. 
Over the years many different solution approaches were proposed. 
\textit{Exact methods} like \textit{branch-and-bound}\cite{lawler1966branch} attempt to find the global optimum of a COP based on smart and efficient ways of searching the solution space. While they are often able to find the optimal solution to small scale COPs in reasonable time, they require a significant amount of time to tackle larger instances of sizes relevant for practical applications. 
For that reason, \textit{heuristic methods} were proposed which usually cannot guarantee to find the global optimum but sometimes can define a lower bound on the performance, which can be achieved at a minimum, and have shown good empirical performance.
One common and well-known heuristic method is \textit{local search} (LS)\cite{aarts2003local}. The main concept of LS is to iteratively explore the search space of candidate solutions in the close neighborhood of the current solution by applying small (local) changes. 
The simplest local search procedure is employing a hill climbing routine which greedily applies local changes that improve the current solution. However, this approach can easily get stuck in bad local optima from which it cannot escape anymore. 
Therefore, LS is commonly used in combination with \textit{meta-heuristics} which enable the procedure to escape from local optima and achieve better final performance.
The meta-heuristics introduced in section \ref{ss:rw_metaheuristics} are well established in the optimization community and have demonstrated very good performance on a plenitude of different COPs.
 
Since a few years however, there is an increasing interest in leveraging methods from the growing field of machine learning (ML) to solve COPs, as recent developments in neural network architectures, involving the Transformer\cite{vaswani2017attention} and Graph Neural Networks (GNNs)\cite{wu2020comprehensive}, have led to significant progress in this domain. 
Most work based on ML is concerned with auto-regressive approaches to construct feasible solutions\cite{vinyals2015pointer,kool2018attention,hu2017solving,zhang2020learning}, but there is also some work which focuses on iterative improvement\cite{chen2019learning,lu2019learning,wu2021learning} or exact solutions\cite{gasse2019exact,nair2020solving} for COPs.
While the existing iterative improvement approaches share some similarities with meta-heuristics and local search, the exact formulation of the methods is often problem specific and misses the generality of the LS framework as well as a clear definition of the 
intervention points which can be used to control the meta-heuristic procedure for LS.

In this work we present a consistent formulation of learned meta-heuristics in local search procedures and show on two representative problems how it can be successfully applied. In our experiments we compare our method to well-known meta-heuristics commonly used with LS at the example of capacitated vehicle routing (CVRP) and job shop scheduling (JSSP). 
The results show that our GNN-based learned meta-heuristic consistently outperforms these methods in terms of speed and final performance. In further experiments we also establish our method in the context of existing ML solution approaches.

\ \\
\noindent
\textbf{Contributions}
\begin{enumerate}
	\item We identify and describe three independent algorithmic aspects
	of local search for combinatorial optimization, each with
	several alternatives, and formalize their sequential selection
	during an iterative search run as Markov Decision Process. 
	
	\item We design a deep graph neural network as policy model for this MDP,
	yielding a learned controller for local search called NeuroLS.
	
	\item We provide ample experimental evidence that NeuroLS outperforms
	both, well-known general purpose local search controllers from
	the Operations Research literature (so called meta-heuristics)
	as well as earlier machine learning-based approaches.
\end{enumerate}

\section{Related Work}
There is a plenitude of approaches and algorithms to tackle combinatorial optimization problems. 
One common heuristic method is \textit{Local Search} (LS)\cite{aarts2003local}. 
In the classical discrete optimization literature it is often embedded into a meta-heuristic procedure to escape local optima.

\subsection{Construction Heuristics} 
Construction algorithms are concerned with finding a first feasible solution for a given COP. They usually start with an empty solution and consecutively assign values to the respective decision variables to construct a full solution. Well-known methods are e.g.\ the \textit{Savings} heuristic\cite{clarke1964scheduling} for vehicle routing problems or \textit{priority dispatching rules} (PDR)\cite{blackstone1982state} for scheduling.
Most improvement and meta-heuristic methods require a feasible initial solution from which they can start to improve.

\subsection{Meta-Heuristics}\label{ss:rw_metaheuristics}
Meta-heuristics are the go-to method for complex discrete optimization problems. 
They effectively balance the exploration of the solution space (escaping local optima) and the exploitation of promising solutions.
There are two major types of methods:\\

\noindent
\textbf{Single Solution Methods} Single solution methods include many well-known approaches
which are used in combination with LS. 
During the search they only maintain a single solution at a time which is iteratively changed and adapted. 
\textit{Simulated Annealing} (SA) is a probabilistic acceptance strategy based on the notion of controlled cooling of materials first proposed in \cite{kirkpatrick1983optimization}. The idea is to also accept solutions which are worse than the best solution found so far to enable exploration of the search space but with a decreasing probability to increasingly focus on exploitation the further the search advances. 
\textit{Iterated Local Search} (ILS)\cite{lourencco2019iterated} alternates between a diversification and an intensification phase. In the diversification step the current solution is perturbed 
while the alternating intensification step executes a greedy local search with a particular neighborhood.
\textit{Variable Neighborhood Search} (VNS)\cite{mladenovic1997variable} employs a similar diversification step but focuses on a systematic control of the LS neighborhood in the intensification phase, by changing the type of the applied LS move frequently after each perturbation in a predefined order.
\textit{Tabu Search} (TS) was proposed by Glover\cite{glover1986future} and is based on the possible acceptance of non-improving moves and a so called tabu list, a kind of filter preventing moves which would keep the search stuck in local optima. This list acts as a memory which in the simplest case stores the solutions of the last $k$ iterations and prevents changes which would move the current solution back to solutions encountered in recent steps. 
Instead of using a tabu list, \textit{Guided Local Search} (GLS)\cite{voudouris2010guided} relies on penalties for different moves to guide the search. These penalties are often based on problem specific features in the solution, e.g.\ the edges between nodes in a routing problem, and are added to the original objective function of the problem when the LS gets stuck. 

\ \\
\noindent
\textbf{Population-based Methods} In comparison to single solution approaches population-based methods maintain a whole pool of different solutions throughout the search. 
Methods include different evolutionary algorithms, particle swarm optimization and other bio-inspired algorithms such as ant colony optimization \cite{gendreau2010handbook}.
Their main idea is based on different adaption, selection and recombination schemes to refine the solution pool during search in order to find better solutions. 

While a learned population based meta-heuristic is interesting and potentially promising, in this paper we focus on the impact and effectiveness of a learned single solution approach. 
More information on advanced meta-heuristics and their extensions can be found in \cite{gendreau2010handbook}.

\subsection{Machine Learning based Methods}\label{ss:rw_ml}
In recent years also many approaches employing ML methods have been proposed. While some work \cite{vinyals2015pointer,joshi2019efficient,thyssens2022supervised} relies on supervised learning, most current state-of-the-art methods use reinforcement learning (RL). 
A large fraction of the work focuses on auto-regressive models which learn to sequentially construct feasible solutions from scratch.
Such methods have been proposed for many common COPs including the TSP\cite{bello2016neural}, CVRP\cite{nazari2018reinforcement,kool2018attention,kwon2020pomo}, CVRP-TW\cite{falkner2020learning} and JSSP\cite{zhang2020learning,park2021learning,hottung2021efficient,park2021schedulenet}. 
The second type of methods is concerned with improvement approaches.
Chen et al.\cite{chen2019learning} design a model to rewrite sub-sequences of the problem solution for the CVRP and JSSP based on a component which selects a specific element of the solution and a second component parameterizing a heuristic move to change that part of the solution. However, their model is limited to specific problem settings with a fixed number of jobs and machines or customers while our approach works seamlessly for different problem sizes, as we show in the experiments in section \ref{ss:results}.
In contrast, the authors in \cite{lu2019learning} learn a policy that selects a specific LS move at each iteration. 
However, their method incurs prohibitively large computation times and for this reason is not competitive with any of the recent related work \cite{kwon2020pomo,kool2021deep,ma2021learning}.
The authors in \cite{hottung2019neural} learn a repair operator to re-construct heuristically destroyed solutions in a Large Neighborhood Search. 
Finally, da Costa et al.\cite{d2020learning} learn a model to select node pairs for 2-opt moves in the TSP while 
Wu et al.\cite{wu2021learning} learn a similar pair-wise selection scheme for 2-opt, node swap and relocation moves in TSP and CVRP.
The authors in \cite{ma2021learning} further improve on the method in \cite{wu2021learning} by introducing the Dual-aspect collaborative Transformer (DACT) model. Although there exist advanced inference approaches\cite{hottung2021efficient} to further improve the performance of auto-regressive ML methods on COPs, here we focus on the vanilla inference via greedy decoding or sampling.

While these methods share some similarities with our approach, they ignore the importance of being able to reject unpromising moves to escape local optima, whereas our approach specifically focuses on this important decision point to effectively control the search. 
Moreover, our approach is also able to learn when to apply a particular perturbation if the rejection of a sequence of moves is not sufficient for exploration.
A detailed overview of the current machine learning approaches to COPs is given in \cite{bengio2021machine,mazyavkina2021reinforcement}.

\section{Preliminaries}
\subsection{Problem Formulation}
A Combinatorial Optimization Problem $\Omega$ is defined on its domain $D_{\Omega}$ which is the set of its instances $x \in D_{\Omega}$. A COP instance $x$ is usually given by a pair ($S_{\Omega}$, $f_{\Omega}$) of the solution space $S$ consisting of all feasible solutions to $\Omega$ and a corresponding cost function $f: S \to \IR$.
Combinatorial optimization problems normally are either to be minimized or maximized. In this paper we consider all COPs to be problems for which the cost of a corresponding objective function has to be minimized. The main concern is to find a solution $s^* \in S$ representing a \textit{global optimum}, i.e.\ $f(s^*) \leq f(s) \ \forall s \in S$.

\subsection{Local Search}
LS is a heuristic search method which is based on the concept of neighborhoods. A neighborhood $\mathcal{N}(s) \subseteq S$ of solution $s \in S$ represents a set of solutions which are somehow close to $s$. This ``closeness'' is defined by the neighborhood function $\mathcal{N}^{\phi}$ w.r.t.\ some problem specific operator $\phi \in \Phi_{\Omega}$ (e.g.\ an exchange of nodes in a routing problem). 
Moreover, we always consider $s$ to be part of its own neighborhood, i.e.\ $s \in \mathcal{N}(s)$. Then the \textit{local optimum} $\hat{s}$ in the neighborhood $\mathcal{N}$ satisfies $f(\hat{s}) \leq f(s) \ \forall s \in \mathcal{N}(\hat{s})$. A general LS procedure (see algorithm \ref{alg:ls}) iterates through the neighborhood $\mathcal{N}(s)$ of the current solution $s$ until it finds the local optimum $\hat{s}$.

\SetKwComment{Comment}{// }{ }
\SetKwFunction{acpt}{accept}
\SetKwFunction{stp}{stop}
\SetKwInOut{Input}{input}
\DontPrintSemicolon
\begin{algorithm}[ht]
	\caption{Local Search}\label{alg:ls}
	\Input{
		cost function $f$, solution $s$, neighborhood function $\mathcal{N}$,\\
		acceptance rule $\acpt$, stopping rule $\stp$, 
	}
	\While{not $\stp(s)$	
	}{	
		find $s' \in \mathcal{N}(s)$ for which $\acpt(s, s')$\\		
		$s \gets s'$
	}
	\Return{s}
\end{algorithm}

\subsection{Meta-Heuristics}
Meta-heuristics wrap an LS procedure to enable it to escape from local optima and to explore the solution space more efficiently. Some of the most common meta-heuristic strategies were described in section \ref{ss:rw_metaheuristics}. Algorithm \ref{alg:mh} describes a general formulation of a single solution meta-heuristic procedure.
Each particular strategy takes different decisions about restarting or perturbing the current solution, configuring the local search and accepting intermediate candidate solutions $s'$ (see algorithm \ref{alg:ls}). Some decisions can be fixed and are treated as hyper-parameters for some methods. For example, an SA procedure normally does not select a specific neighborhood but just decides about acceptance during the LS. Other approaches like VNS greedily accept all improving moves but apply a perturbation and select a new operator neighborhood every time a local optimum has been reached.

\SetKwComment{Comment}{// }{ }
\SetKwFunction{constr}{construct}
\SetKwFunction{ls}{LocalSearch}
\SetKwFunction{nbh}{GetNeighborhood}
\SetKwFunction{ops}{GetOperator}
\SetKwFunction{acpt}{accept}
\SetKwFunction{stp}{stop}
\SetKwFunction{rest}{restart}
\SetKwFunction{pert}{perturb}
\SetKwInOut{Input}{input}
\DontPrintSemicolon
\begin{algorithm}[ht]
	\caption{Meta-Heuristic (Single Solution)}\label{alg:mh}
	
	\Input{Solution space $S$, cost function $f$, stopping criterion} 
	
	$s \gets \constr(S)$	\Comment*[r]{Construct initial solution}
	\While{not stopping criterion}{	
		$s \gets \pert(S, s)$	\Comment*[r]{Decide if to perturb/restart}
		$\mathcal{N} \gets \nbh(S, s)$   	\Comment*[r]{Define search neighborhood}
		$s \gets \ls(f, s, \mathcal{N}, \acpt, \stp)$	\Comment*[r]{Execute local search}
	}
	\Return{s}
\end{algorithm}

\section{Proposed Method}
\subsection{Intervention Points of Meta-Heuristics for Local Search}
The application of meta-heuristic strategies to an underlying LS involves several points of intervention, at which decisions can be done to help the search escape local optima in order to guide it towards good solutions. In the following, we define three such intervention points that have a significant impact on the search:

\begin{enumerate}
	\item \textit{Acceptance}: 
	The first intervention point is the acceptance of candidate solutions $s'$ in a local search step (algorithm \ref{alg:ls}, line 2). A simple hill climbing heuristic is completely greedy and only accepts improving moves, which often leaves the search stuck in local optima very quickly. In contrast, other approaches like SA will also accept non-improving moves with some probability. 
	\item \textit{Neighborhood}: 
	The second possible decision a meta-heuristic can make is the selection of a particular operator $\phi$ that defines the neighborhood function $\mathcal{N}^{\phi}$ for the LS (algorithm \ref{alg:mh}, line 4). Possible operators for scheduling or routing problems could for example be a node exchange. While many standard approaches like SA and ILS only use one particular neighborhood which they treat as a hyper-parameter, VNS is an example for a method that selects a different operator at each step.
	\item \textit{Perturbation}: 
	Many meta-heuristics like ILS and VNS employ perturbations $\psi \in \Psi_{\Omega}$ to the current solution to move to different regions of the search space and escape particularly persistent local optima. Such a perturbation can simply be a restart from a new stochastically constructed initial solution, a random permutation of (a part of) the current solution or a random sequence of node exchanges.
	The decision \textit{when} to employ such a perturbation (algorithm \ref{alg:mh}, line 5) is usually
	done w.r.t.\ a specific pre-defined number of steps without improvement.

\end{enumerate}

\subsection{Meta-Heuristics as Markov Decision Process}\label{ss_mh_mdp}
In this section we formulate meta-heuristics in terms of an MDP\cite{sutton2018reinforcement} to enable the use of RL approaches to learn a parameterized policy to replace them. In general an MDP is given by a tuple 
$(\mathcal{S}, \mathcal{A}, \mathcal{P}(s_t, a_t), \mathcal{R}(s_t, a_t))$ representing the set of states $\mathcal{S}$, the set of actions $\mathcal{A}$, a transition probability function $\mathcal{P}(s_t, a_t)$ and a reward function $\mathcal{R}(s_t, a_t)$. For our method we define these entities in terms of meta-heuristic decisions as follows:

\ \\
\noindent
\textbf{States} \ 
We define the state $s_t$ of the problem at time step $t$ with slight abuse of notation as the solution $s$ at time step $t$, combined with 1) its cost $f(s)$, 2) the cost $f(\hat{s}_t)$ of the best solution found so far, 3) the last acceptance decision, 4) the last operator used, 5) current time step $t$, 6) number of LS steps without improvement and 7) the number of perturbations or restarts.

\ \\
\noindent
\textbf{Actions} \ 
Depending on the policy we want to train, we define the action set as the combinatorial space of
\begin{enumerate}
	\item \textit{Acceptance} decisions: a boolean decision variable of either accepting or rejecting the last LS step
	\begin{equation}
		\mathcal{A}_{\text{A}} \coloneqq \{0, 1\},
	\end{equation}
	
	\item \textit{Acceptance}-\textit{Neighborhood} decisions: the joint space of the acceptance of the last move and the set of possible operators $\phi \in \Phi$ to apply in the next step
	\begin{equation}
		\mathcal{A}_{\text{AN}} \coloneqq \{0, 1\} \times \Phi,
	\end{equation}
	
	\item \textit{Acceptance}-\textit{Neighborhood}-\textit{Perturbation} decisions: the joint space of acceptance and the combined sets of operators $\phi \in \Phi$ and perturbations $\psi \in \Psi$
	\begin{equation}
		\mathcal{A}_{\text{ANP}} \coloneqq \{0, 1\} \times \{\Phi \cup \Psi\}.
	\end{equation}
\end{enumerate}

\ \\
\noindent
\textbf{Transitions} \ 
The transition probability function $\mathcal{P}(s_t, a_t)$ models the state transition from state $s_t$ to the next state $s_{t+1}$ depending on action $a_t$ representing the acceptance decision, the next operator $\phi$ 
and a possible perturbation or restart (part of the problem state or action in case of $\mathcal{A}_{\text{ANP}}$).

\ \\
\noindent
\textbf{Rewards} \ 
The reward function $\mathcal{R}(s_t, a_t)$ defines the reward for a transition from state $s_t$ to the next state $s_{t+1}$.
Here we define the reward $r_t$ of the transition as the relative improvement of the last LS step (defined by action $a_t$) w.r.t.\ the cost of the best solution found until $t$ and clamped at 0 to avoid negative rewards:
\begin{equation}
	r_t \coloneqq \max(f(\hat{s}_t) - f(s_{t+1}), 0)
\end{equation}

\ \\
\noindent
\textbf{Policy} \ 
We employ Deep Q-Learning\cite{van2016deep} and parameterize the learned policy $\pi_{\theta}$ via a softmax over the corresponding Q-function $Q_{\theta}(s_t, a_t)$ which is represented in turn by our GNN-based encoder-decoder model with trainable parameters $\theta$:
\begin{equation}
	\pi_{\theta}(a_t \mid s_t) = \frac{\exp(Q_{\theta}(s_t, a_t))}{\sum_{\mathcal{A}} \exp(Q_{\theta}(s_t, a_t))}.
\end{equation}

\subsection{Model Architecture}
In this section we describe our encoder and decoder models to parameterize $Q_{\theta}(s_t, a_t)$. 
Many COPs like routing and scheduling problems have an underlying graph structure which can be used when encoding these problems and that provides an effective inductive bias for the corresponding learning methods. 

In general we assume a graph $\mathcal{G} = (\mathcal{V}, \mathcal{E})$ with the set of nodes $\mathcal{V}$, $N = |\mathcal{V}|$ and the set of directed edges $\mathcal{E} \subseteq \{(i, j) \subseteq \mathcal{V} \}$. 
Moreover, we assume an original node feature matrix $X \in \IR^{N \times d_{\text{in}}}$ and edge features $e_{i, j} \in \IR$ for each edge $(i,j)$. 
To leverage this structural information many authors have used Recurrent Neural Networks (RNN)\cite{vinyals2015pointer,bello2016neural,chen2019learning}, Transformers\cite{vaswani2017attention,kool2018attention,kwon2020pomo,ma2021learning} or Graph Neural Networks (GNN)\cite{joshi2019efficient,d2020learning,zhang2020learning}. 

\ \\
\noindent
\textbf{Encoder} \\
For our model we employ the 1-GNN operator of \cite{morris2019weisfeiler} which is able to work with edge weights if they are present. One such layer 
is defined as
\begin{align}
	h^{(l)}_i &= \GNN^{(l)}(h^{(l-1)}_i) \nonumber \\
			  &= \sigma\big( \MLP^{(l)}_1(h^{(l-1)}_i) + \MLP^{(l)}_2(\sum_{j\in \mathcal{H}(i)} e_{j, i} \cdot h^{(l-1)}_j) \big),
\end{align}
where $h^{(l-1)}_i \in \IR^{1 \times d_{\text{emb}}}$ is the latent feature embedding of node $i$ at the previous layer $l-1$, $\mathcal{H}(i)$ is the 1-hop graph neighborhood of node $i$, $\MLP^{(l)}_1$ and $\MLP^{(l)}_2$ are Multi-Layer Perceptrons $\MLP: \IR^{d_{\text{emb}}} \to \IR^{d_{\text{emb}}}$ and $\sigma()$ is a suitable element-wise non-linearity, for which we use GeLU\cite{hendrycks2016gaussian}. 
Furthermore, we add residual connections and layer normalization\cite{ba2016layer} to each layer.

In the first layer the latent feature vector $h^{(0)}_i$ is created by feeding the original node features $x_i$ into an $\MLP: \IR^{d_{\text{in}}} \to \IR^{d_{\text{emb}}}$: 
\begin{equation}
	h^{(0)}_i = \MLP^{(0)}(x_i).
\end{equation}
and another $\MLP^{(L)}: \IR^{d_{\text{emb}}} \to \IR^{d_{\text{emb}}}$ is placed at the end of the GNN stack.

\begin{figure}[t!]
	\centering
	\includegraphics[width=\textwidth]{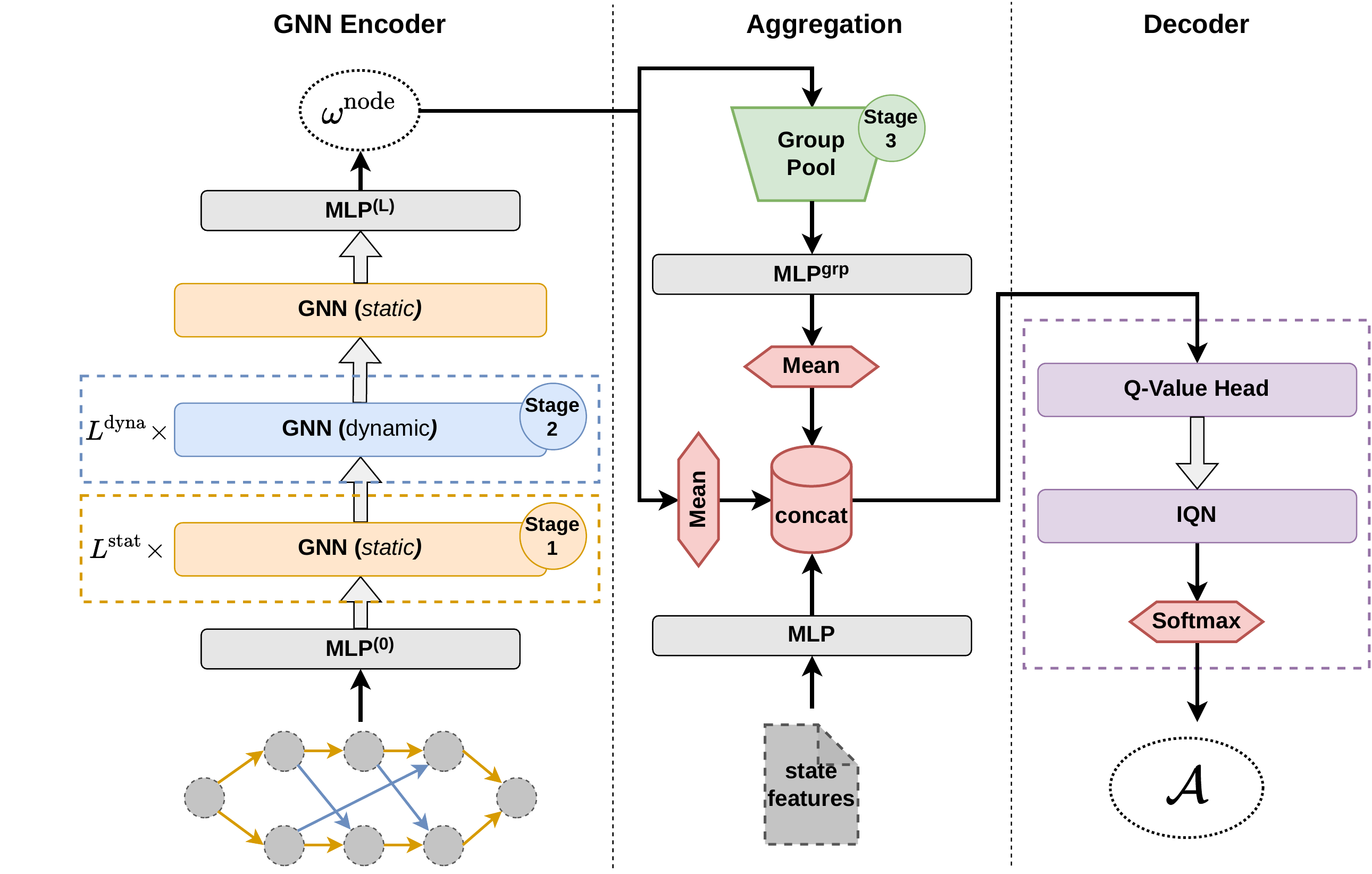}
	\caption{
		Visualization of the NeuroLS model architecture. 
	} 
	\label{fig:arch}
\end{figure}

In order to further leverage structural information, we introduce 3 stages to compute the latent embeddings. The first stage uses the edge set $\mathcal{E}^{\text{stat}}$ of the \textit{static} node neighborhood graph, e.g.\ the $K$ nearest neighbors graph for the CVRP or the Directed Acyclic Graph (DAG) representing the order of operations for each job in scheduling problems. This edge set is fixed and does not change throughout the search. 
In contrast, the second stage utilizes the edge set $\mathcal{E}^{\text{dyna}}$ representing e.g.\ the tours in routing or the machine graph in scheduling, which is \textit{dynamic} and usually changes in every LS step.
Our proposed network architecture consists of $L^{\text{stat}}$ GNN layers for the static graph, followed by $L^{\text{dyna}}$ layers which propagate over the dynamic graph. Finally, we add another layer, again using the static edges, to consolidate the dynamic information over the static graph, leading to a total of $L = L^{\text{stat}} + L^{\text{dyna}} +1$ GNN layers.

The final stage serves to refine the embedding via aggregation based on the dynamic information of group membership which is present in the solution. Each node normally belongs to one of $K$ (not necessarily disjoint) groups $\mathcal{M}_k$ of the solution, e.g.\ to a particular tour or machine. Following this idea we pool the final embeddings $\omega^{\text{node}}_i$ from the GNN stack w.r.t.\ their membership and feed them through another MLP:
\begin{equation}
	\omega^{\text{grp}}_k = \MLP^{\text{grp}}\big([\MAX(\omega^{\text{node}}_i \mid i \in \mathcal{M}_k); \MEAN(\omega^{\text{node}}_i \mid i \in \mathcal{M}_k)]\big),
\end{equation}
with max and mean pooling $\IR^{N \times d_{\text{emb}}} \to \IR^{K \times d_{\text{emb}}}, K << N$ over the node dimension $N$, $\mathcal{M}_k$ as membership to the $k$-th group ($k$-th tour, $k$-th machine, etc.) and $[\ ;]$ representing concatenation in the embedding dimension $d_{\text{emb}}$.

Finally, the additional features of the state representation (current cost, best cost, last acceptance, etc.) described in the last section, are concatenated and projected by a simple linear layer to create an additional latent feature vector $\omega^{\text{feat}} \in \IR^{d_{\text{emb}}}$.

\ \\
\noindent
\textbf{Decoder} \\
Our decoder takes the different embeddings created in the encoder, aggregates the node embeddings $\omega^{\text{node}} \in \IR^{N \times d_{\text{emb}}}$ and group embeddings $\omega^{\text{grp}} \in \IR^{K \times d_{\text{emb}}}$ via a simple mean over the node and group dimension and concatenates them with the feature embedding $\omega^{\text{feat}} \in \IR^{d_{\text{emb}}}$. 
This representation is the input to a final 2-layer MLP regression head $\IR^{3*d_{\text{emb}}} \to \IR^{|\mathcal{A}|}$ which outputs the value predictions of the Q-function. The full architecture is shown in figure \ref{fig:arch}.

\subsection{Reinforcement Learning Algorithm}
To train our policy model we employ Double Deep Q-Learning\cite{van2016deep} with $n$-step returns\cite{sutton2018reinforcement}
and Implicit Quantile Networks (IQN)\cite{dabney2018implicit}. IQNs enable a distributional formulation of Q-Learning where the Deep Q-Network is trained w.r.t.\ an underlying value distribution represented by a learned quantile function instead of single point estimates. In order to represent this learned quantile function IQNs are introduced as a small additional neural network which is jointly trained to transform samples from a base distribution (e.g.\ uniform) to the respective quantile values of the target distribution, i.e.\ the distribution over the returns. Further details can be found in appendix \ref{appx:rl}.

\section{Experiments}
\subsection{Applications}
\noindent
\textbf{Job Shop Scheduling Problem (JSSP)}
The JSSP is concerned with scheduling a number of jobs $J$ on a set of $K$ machines denoted by $M$. Each job consists of a sequence of operations $O_{ij}$ with fixed processing times $p_{ij}$ which need to be processed in a predefined order. In the simplest problem variant every job has exactly one operation on each machine. A solution to the problem consists of the exact order of the operations on all machines.
In this paper we choose to minimize the \textit{makespan}, which is the longest time span from start of the first operation until the end of the last one to finish, corresponding to the longest path in the respective DAG representation of the problem. In the following we denote the size of a JSSP instances as $|J|\times |M|$. We follow \cite{zhang2020learning} in creating instances for training and validation by sampling processing times from a uniform distribution.

\ \\
\noindent
\textbf{Capacitated Vehicle Routing Problem (CVRP)}
The CVRP consists of a set of $N$ customer nodes and a depot node. It is concerned with serving the demands $q_i$ of the customer nodes in tours starting and ending at the depot node by employing $K$ homogeneous vehicles with a fixed capacity $Q > 0$. The tour of vehicle $k \in K$ is a sequence of indices w.r.t.\ a subset of all customers nodes representing the order in which vehicle $k$ visits the respective nodes. A set of feasible tours serving all customer nodes represents a solution to the problem, whereas the objective is to minimize the total length of all tours. We follow \cite{kool2018attention} in creating the training and validation sets by generating instances with coordinates uniformly sampled in the unit square.

\subsection{Setup}
For the JSSP we implement a custom LS solver in python. It implements four different node moves and a perturbation operator based on a sequence of such moves. As construction heuristic and for restarts we implement several stochastic variants of common PDRs (see appendix \ref{appx:jssp_solver} for more details).
The LS for the CVRP uses the C++ based open source solver VRPH\cite{groer2010library} for which we implement a custom wrapper and interface to expose and support the necessary intervention points. 
VRPH includes several different LS moves and perturbation operators including 2-opt, 3-opt and different exchange and point moves. 
We provide our model code and that of our JSSP solver in our github repository\footnote{\url{https://github.com/jokofa/NeuroLS}}.

Preliminary experiments showed that the CET and 2-opt moves performed best for the JSSP and CVRP respectively, when used for meta-heuristics which do not select the operator. Thus, we employ these operators in our main experiments. 
We train all our models for 80 epochs with 19200 transitions each and pick the model checkpoint with the best validation performance.

Hyperparameters for NeuroLS and all meta-heuristics are tuned on a validation set consisting of 512 generated random instances. This is in contrast to most classical approaches which fine tune their hyper-parameters directly on the benchmark dataset. We argue that our approach facilitates a better and more objective comparison between these methods. Furthermore, we fix the random seed for all experiments.

We train 3 different types of policies for NeuroLS, one for each of the action spaces described in section \ref{ss_mh_mdp}. In the experiments we denote these policies as NLS$_{\text{A}}$, NLS$_{\text{AN}}$ and NLS$_{\text{ANP}}$.
As analyzed in \cite{wu2021learning}, random solutions do not provide a good starting point for improvement approaches. For that reason we initialize the solutions of NeuroLS and the meta-heuristics with the FDD/MWKR PDR\cite{sels2012comparison} for scheduling and the well known savings construction heuristic\cite{clarke1964scheduling} for the CVRP. 
Further details on training and hyper-parameters can be found in Appendix \ref{appx:training}.

\subsection{Results}\label{ss:results}

\begin{table*}[th]
	\caption{
		Results of state-of-the-art machine learning based construction methods and local search approaches (100 iterations) on the Taillard benchmark\cite{taillard1993benchmarks}. For instances of size 50x15, 50x20 and 100x20 we use the NeuroLS model trained on instances of size 30x15 and 30x20 respectively. Percentages are the average gap to the best known upper bound. Best gap is marked in \textbf{bold}.
	}
	\label{tab:results_jssp}
	\centering
	\begin{tabular}{l|rrrrr|rrr|r}
		\toprule
		\textbf{Model} & \multicolumn{8}{c|}{\textbf{Instance size}} & \textbf{Avg} \\
		\hline
		& 15x15 & 20x15 & 20x20 & 30x15 & 30x20 & 50x15 & 50x20 & 100x20 &  \\
		\midrule
		\multicolumn{10}{c}{\textbf{ML-based}} \\
		L2D\cite{zhang2020learning}	&	25.92\%	&	30.03\%	&	31.58\%	&	32.88\%	&	33.64\%	&	22.35\%	&	26.37\%	&	13.64\% & 27.05\%	\\
		L2S\cite{park2021learning} 	&	20.12\%	&	24.83\%	&	29.25\%	&	24.59\%	&	31.91\%	&	15.89\%	&	21.39\%	&	9.26\% & 22.16\%	\\
		SN\cite{park2021schedulenet}	&	15.32\%	&	19.43\%	&	17.23\%	&	18.95\%	&	23.75\%	&	13.83\%	&	13.56\%	&	6.67\% & 16.09\%	\\
		\midrule
		\multicolumn{10}{c}{\textbf{Meta-heuristic + Local Search}} \\
		SA		&	13.92\%	&	17.01\%	&	17.16\%	&	17.53\%	&	21.59\%	&	12.50\%	&	13.11\%	&	6.61\% & 14.93\%	\\
		SA$_{\text{restart}}$					  &	13.77\%	&	17.01\%	&	17.57\%	&	17.62\%	&	21.78\%	&	12.54\%	&	13.22\%	&	6.75\% & 15.03\%	\\
		ILS		&	11.57\%	&	13.57\%	&	13.85\%	&	16.07\%	&	18.72\%	&	12.65\%	&	12.15\%	&	6.72\% & 13.16\%	\\
		ILS+SA	&	13.32\%	&	16.05\%	&	15.38\%	&	16.93\%	&	19.74\%	&	13.07\%	&	13.43\%	&	7.08\% & 14.37\%	\\
		VNS		&	9.96\%	&	13.71\%	&	14.51\%	&	15.77\%	&	18.69\%	&	11.64\%	&	11.92\%	&	6.26\% & 12.81\%	\\
		\midrule
		\multicolumn{10}{c}{\textbf{NeuroLS}} \\
		NLS$_{\text{A}}$	&	\textbf{9.76}\%	&	13.33\%	&	13.02\%	&	15.29\%	&	17.94\%	&	11.81\%	&	11.96\%	&	6.33\% & 12.43\%	\\
		NLS$_{\text{AN}}$	&	10.32\%	&	\textbf{13.18}\%	&	\textbf{12.95}\%	&	\textbf{14.91}\%	&	17.78\%	&	11.87\%	&	12.02\%	&	6.22\% & \textbf{12.41}\%	\\
		NLS$_{\text{ANP}}$	&	10.49\%	&	16.32\%	&	15.24\%	&	15.35\%	&	\textbf{17.64}\%	&	\textbf{11.62}\%	&	\textbf{11.76}\%	&	\textbf{6.09}\% & 13.06\%	\\
		\bottomrule
	\end{tabular}
\end{table*}

\begin{table*}[th]
	\caption{
		Results of state-of-the-art machine learning based methods and local search approaches (200 iterations) on the Uchoa benchmark\cite{uchoa2017new}. For all instances sizes we use the NeuroLS model trained on instances of size 100. Percentages are the average gap to the best known solution. Best gap is marked in \textbf{bold}.
	}
	\label{tab:results_cvrp}
	\centering
	\begin{tabular}{l|rrrrrrrr|r}
		\toprule
		\textbf{Model} & \multicolumn{8}{c|}{\textbf{Instance Group}} & \textbf{Avg} \\
		\hline
		& \multicolumn{2}{c}{\textit{n100}} & \multicolumn{2}{c}{\textit{n150}} & \multicolumn{2}{c}{\textit{n200}} & \multicolumn{2}{c|}{\textit{n250}} &   \\
		& cost & time & cost & time & cost & time & cost & time & \\
		\midrule
		\multicolumn{10}{c}{\textbf{ML-based}} \\
		POMO\cite{kwon2020pomo} &	 17.24\%	&	0.3 	&	 12.81\%	&	0.4	&	 20.52\%	&	0.4 	&	 15.14\%	&	 0.4	&	 16.43\%	\\
		POMO\cite{kwon2020pomo} (aug) &	 6.32\%	&	\textbf{0.1} 	&	9.41\% 	&	\textbf{0.1}	&	13.47\% 	&	\textbf{0.1} 	&	 9.21\%	&	\textbf{0.1} 	&	9.60\% 	\\
		DACT\cite{ma2021learning}	&	13.19\% 	&	15.1 	&	20.26\% 	&	21.2	&	16.91\% 	&	27.2 	&	24.93\% 	&	 33	&	18.82\% 	\\
		DACT\cite{ma2021learning} (aug) &	 11.52\%	&	16.6 	&	17.75\% 	&	23.1	&	15.34\% 	&	29.8 	&	 21.86\%	&	37.8 	&	16.62\% 	\\
		\midrule
		\multicolumn{10}{c}{\textbf{Meta-heuristic + Local Search}} \\
		ORT\cite{ortools} GLS	&	6.91\% 	&	9.4 	&	 10.46\%	&	10.9	&	 \textbf{7.80}\%	&	13.4 	&	12.12\% 	&	 5.0	&	 9.32\%	\\
		ORT\cite{ortools} TS	&	6.78\% 	&	 39.0	&	 10.55\%	&	109.5	&	 7.85\%	&	126.6 	&	12.13\% 	&	7.0 	&	 9.33\%	\\
		SA		&	6.92\% 	&	0.7 	&	5.79\% 	&	1.3	&	 16.63\%	&	2.1 	&	5.92\% 	&	3.3 	&	 8.81\%	\\
		SA$_{\text{restart}}$	&	 6.96\%	&	 0.7	&	5.79\% 	&	1.3	&	16.67\% 	&	2.0 	&	5.99\% 	&	 3.0	&	8.85\% 	\\
		ILS		&	 6.56\%	&	0.8 	&	5.96\% 	&	1.5	&	16.73\% 	&	 2.4	&	6.08\% 	&	3.7 	&	 8.83\%	\\
		ILS+SA	&	 6.94\%	&	0.6 	&	6.01\% 	&	1.2	&	 16.72\%	&	 1.9	&	6.04\% 	&	2.8 	&	 8.93\%	\\
		VNS		&	 7.99\%	&	0.4 	&	 6.55\%	&	0.7	&	17.27\% 	&	 0.9	&	6.36\% 	&	1.4 	&	 9.54\%	\\
		\midrule
		\multicolumn{10}{c}{\textbf{NeuroLS}} \\
		NLS$_{\text{A}}$ 	&	 5.43\%	&	1.5 	&	5.23\% 	&	2.4	&	15.97\% 	&	3.6 	&	5.22\% 	&	5.3 	&	7.96\% 	\\
		NLS$_{\text{AN}}$	&	\textbf{5.42}\% 	&	1.7 	&	\textbf{4.90}\% 	&	2.7	&	15.85\% 	&	4.0 	&	 \textbf{5.08}\%	&	5.9 	&	 \textbf{7.81}\%	\\
		NLS$_{\text{ANP}}$	&	 \textbf{5.42}\%	&	1.7 	&	\textbf{4.90}\% 	&	2.7	&	15.85\% 	&	4.0 	&	 \textbf{5.08}\%	&	6.1 	&	 \textbf{7.81}\%	\\
	
		\bottomrule
	\end{tabular}
\end{table*}

\ \\
\noindent
\textbf{JSSP} We evaluate all methods on the well-known benchmark dataset of Taillard\cite{taillard1993benchmarks}. It consists of 80 instances of size 15x15 up to 100x20. We compare our model against common meta-heuristic baselines including SA, SA with restarts, ILS, ILS with SA acceptance and VNS. Moreover, we report the results of three recent state-of-the-art ML-based approaches: \textit{Learning to dispatch} (L2D)\cite{zhang2020learning}, \textit{Learning to schedule} (L2S)\cite{park2021learning} and \textit{ScheduleNet} (SN)\cite{park2021schedulenet}.
We follow \cite{zhang2020learning} in training different models for problem sizes 15x15 up to 30x20 and apply the 30x15 and 30x20 model to larger instances of size 50x15, 50x20 and 100x20 to evaluate its generalization capacity. All models are trained for 100 LS iterations but evaluated for 50-200. The aggregated results per group of same size are shown in table \ref{tab:results_jssp} (results per instance can be found in appendix \ref{appx:benchmark}). Results are reported as percentage gaps to the best known solution.

First of all, the results show that NeuroLS is able to outperform all other meta-heuristics on all sizes of instances. The different policies differ in how well they work for different problem sizes. While NLS$_{\text{A}}$ works best for the smallest 15x15 instances, it is outperformed by NLS$_{\text{AN}}$ on medium sized instances and NLS$_{\text{ANP}}$ achieves the best results for large instances. This is to some extent expected, since the effect of specific LS operators and more precise perturbations is greater for larger instances while this does not seem to be necessary for rather small instances.
VNS is the best of the meta-heuristic approaches which is able to beat NLS$_{\text{ANP}}$ on the smaller instances and NLS$_{\text{A}}$ and NLS$_{\text{AN}}$ at least on some of the larger ones.

In general, the iterative methods based on LS achieve much better results than the ML-based auto-regressive methods, which can be seen by the large improvements that can be achieved in just 100 iterations, reducing the gap by an average of 3.7\% compared to the best ML-method SN\cite{park2021schedulenet}.

\ \\
\noindent
\textbf{CVRP} For the CVRP we use the recent benchmark dataset of Uchoa et al.\cite{uchoa2017new} and select all instance up to 300 customer nodes. We define four groups of instances with 100-149 nodes as \textit{n100}, 150-199 nodes as \textit{n150}, 200-249 as \textit{n200} and 250-299 as \textit{n250}. We compare against the same meta-heuristics mentioned above and additionally to GLS and TS provided by the OR-Tools (ORT) library\cite{ortools}. Furthermore, we compare to the recent state-of-the-art ML approaches POMO\cite{kwon2020pomo} and DACT\cite{ma2021learning} which outperformed all other ML methods mentioned in section \ref{ss:rw_ml} in their experiments and provide open source code, which is why we consider them to be sufficient for a suitable comparison.

Since most ML-based methods (POMO, DACT, etc.) do not respect the maximum vehicle constraint for all instances of the Uchoa benchmark, we follow\cite{ma2021learning} in removing this constraint and treat the benchmark dataset as a highly diverse test set w.r.t.\ the distributions of customers and number of required vehicles. 
This is also consistent with the general goal of the ML-based methods, which is not to achieve the best known results but to find sufficiently good results in reasonable time. In this case we control the computational resources spent on the search by specifying a particular number of iterations for the search. Furthermore, we evaluate all models with a batch size of 1. 

The results presented in table \ref{tab:results_cvrp} show that NeuroLS is also able to outperform the meta-heuristic approaches mentioned above on all instance groups. Moreover, our approach also outperforms the best state-of-the-art ML-based methods on all groups but \textit{n200}, where \cite{kwon2020pomo} and \cite{ma2021learning} with additional instance augmentations (aug) outperform NeuroLS by a small margin. The OR-Tools implementation of GLS and TS outperforms our method only on the \textit{n200} instances, although they require prohibitively large runtimes (wall-clock time) several magnitudes larger than our approach. In terms of runtimes we also outperform DACT\cite{ma2021learning} by a magnitude, while the learned auto-regressive construction method POMO\cite{kwon2020pomo} is the fastest over all.

Finally, the experiment results also show that our method is well able to generalize to problem sizes as well as numbers of iterations unseen in training. We show this on the JSSP instances of size 50x15, 50x20 and 100x20 and basically on all Uchoa instances for the CVRP, which are larger than the 100 node instances used during training.

\section{Conclusion}
In this paper we identify three important intervention points in meta-heuristics for LS on COPs and incorporate them in a Markov Decision Process. We then design a GNN-based controller model which is trained with RL to parameterize three different types of learned meta-heuristics. The resulting methods learn to control the local search by deciding about acceptance, neighborhood selection and perturbations. In comprehensive experiments on two common COPs in scheduling and vehicle routing, NeuroLS outperforms several well-known meta-heuristics as well as state-of-the-art ML-based approaches, confirming the efficacy of our method. 

For future work we consider to replace the problem graph representation with a graph representation of the corresponding local search graph, in which every node represents a feasible solution together with its respective cost. A Curriculum learning strategy (cp. \cite{ma2021learning}) could also yield further improvements regarding performance and generalization.


\bibliographystyle{splncs04}
\bibliography{references}


\newpage
\renewcommand*{\thesubsection}{\Alph{subsection}}
\section*{Appendices}

\subsection{Reinforcement Learning Algorithm}\label{appx:rl}

Let $\pi$ be a policy, then the discounted sum of future rewards is given by the random variable
\begin{equation}
	Z^{\pi}(s, a) = \sum_{t=0}^{\infty} \gamma^t R(s_t, a_t)
\end{equation}
with $s_t \sim P(s_{t-1}, a_{t+1})$. Then the action-value function $Q(s, a) = \mathbb{E}[Z^{\pi}(s, a)]$ can be written in terms of the Bellman equation as
\begin{equation}
	Q^{\pi}(s_t, a_t) = \mathbb{E}[R(s_t, a_t)] + \gamma \mathbb{E}[Q^{\pi}(s_{t+1}, a_{t+1})].
\end{equation}
Furthermore, the real action-value function can be approximated by a parameterized value estimator $Q_{\theta}(s, a)$ which is trained via minimization of the squared error of the temporal difference defined as

\begin{equation}
	\text{TDE} = \left[ r_t + \gamma \max_{a' \in \mathcal{A}} Q_{\theta}(s_{t+1}, a') - Q_{\theta}(s_t, a_t)  \right]^2
\end{equation}
based on an $\epsilon$-greedy policy.

Distributional RL\cite{bellemare} aims to replace the scalar value function $Q^{\pi}$ with a distribution over the returns $Z^{\pi}$. 
For the distributional case there exist analogous formulations for the Bellman equation and the corresponding distributional Bellman optimality operator. 
Implicit Quantile Networks\cite{dabney} are used to transform random samples from a base distribution to the respective quantile values of a target distribution, in this case the distribution over the returns. It is formulated as a risk-sensitive greedy policy (w.r.t.\ a distortion measure $\beta$) represented by a weighted sum over the quantiles:
\begin{equation}
	\pi_{\beta}(s) = \argmax_{a\in \mathcal{A}} Q_{\beta}(s, a).
\end{equation}

Then, assuming two samples $x, x' \sim \mathcal{U}([0, 1])$, the respective TD error is defined as
\begin{equation}
	\text{TDE}_{x, x'} = r_t + \gamma  Z_{x'}(s_{t+1}, \pi_{\beta}(s_{t+1})) - Z_{x}(s_t, a_t)
\end{equation}
and the corresponding loss is estimated over multiple i.i.d. samples $x_i, x'_j \sim \mathcal{U}([0, 1])$.

\subsection{Training and Inference Setup}\label{appx:training}

For training we use the Adam optimizer\cite{kingma} with a learning rate of 0.0005. 
We use $\epsilon=0.95$ for our $\epsilon$-greedy policy which we discount over the course of training to a final value of $\epsilon=0.05$. During inference we set $\epsilon=0.0$. Moreover, we use $n$-step returns with $n=3$ and a discount factor $\gamma=0.99$. The target network of our DQN is updated ever 500 steps. We use a replay buffer of size 32000 and prioritized experience replay\cite{schaul} with $\alpha=0.6$ and $\beta=0.4$. 

All MLPs in our model have a hidden dimension of 128 and we set the embedding dimension of all components to $d_{\text{emb}}=128$. The IQN has a hidden dimension of 256. In the GNN encoder we use $L^{\text{stat}}=3$ and $L^{\text{dyna}}=2$ layers. 

We train all our models for 80 epochs with 19200 transitions each. All experiments are run on an i7-7700K CPU (4.20GHz) and a NVIDIA GeForce 1080TI GPU.

Moreover, all our networks as well as the ML-based baselines are implemented in PyTorch\cite{paszke} and we use the tianshou library\cite{weng} for an efficient RL implementation.

The meta-heuristics use different hyper-parameter configurations which where tuned on the validation set consisting of 512 generated instances. 
The exact configurations for all methods can be found in the configuration files in our github repository.

\subsection{JSSP Solver}\label{appx:jssp_solver}
To solve the JSSP we implement a local search solver in python. We implement several different swap and insertion heuristics to provide a suitable number of LS operators and corresponding neighborhoods. The first one is the CT operator proposed by \cite{laarhoven} which tries to swap adjacent nodes in critical blocks. Next, we have the CET\cite{nowicki} and ECET\cite{kuhpfahl} moves which respectively either swap only nodes at the start or the end of a critical block or both simultaneously. The final implemented operator is the CEI move introduced by \cite{balas}, which shifts a node in a critical block to a new position within the same block. \cite{kuhpfahl} provide a nice visualization and explanation of how all these operators work.

In order to quickly evaluate how good some moves are, without applying them and completely recalculating the resulting makespan, we use the potential estimation strategies by \cite{taillard} and \cite{murovec} which are computed in $O(m)$ where $m$ is the number of nodes between the initial and the final position of the moved nodes. Finally, our perturbation operator is a sequence of random CT moves without any evaluation of potentials, somehow similar to the shaking operator in \cite{sevkli}.
The initial solution for our iterative solver is created by the FDD/MWKR priority dispatching rule introduced in \cite{sels}, which we implement as a stochastic variant for restarting as proposed in \cite{lourenco}.
We will provide the implementation of our solver together with our model code.

\subsection{Additional Results}\label{appx:results}
We provide more detailed results for different methods and several different numbers of iterations in table \ref{appx:tab:results_jssp} for the JSSP and table \ref{appx:tab:results_cvrp} for the CVRP. 

\begin{table*}[th]
	\caption{
		Results of PDRs and local search methods running for 50/100/200 iterations on the Taillard benchmark. For instance sizes 50x15, 50x20 and 100x20 we use the NeuroLS model trained on instances of size 30x15 and 30x20 respectively. Percentages are the gap to the best known upper bound. Best gap is marked in \textbf{bold}.
	}
	\label{appx:tab:results_jssp}
	\centering
\begin{tabular}{l|ccccc|ccc|r}
		\toprule
		\textbf{Model} & \multicolumn{8}{c|}{\textbf{Instance size}} & \textbf{Avg} \\
		\hline
		 & 15x15 & 20x15 & 20x20 & 30x15 & 30x20 & 50x15 & 50x20 & 100x20 &  \\
		\midrule
		\multicolumn{9}{c}{\textbf{PDRs}} \\
		RND 	&	28.90\%	&	36.81\%	&	31.71\%	&	34.97\%	&	36.30\%	&	22.86\%	&	28.05\%	&	16.47\% & 29.51\%	\\
		FIFO	&	23.92\%	&	31.50\%	&	27.42\%	&	31.20\%	&	32.14\%	&	21.15\%	&	23.62\%	&	13.06\% & 25.50\%	\\
		SPT		&	27.14\%	&	32.92\%	&	25.30\%	&	34.54\%	&	34.59\%	&	23.81\%	&	24.96\%	&	15.39\% & 27.33\%	\\
		MWKR	&	18.72\%	&	23.94\%	&	21.52\%	&	23.91\%	&	24.97\%	&	16.76\%	&	17.54\%	&	8.33\% & 19.46\%	\\
		MOPNR	&	20.54\%	&	23.99\%	&	21.60\%	&	22.44\%	&	25.10\%	&	17.32\%	&	17.96\%	&	9.24\% & 19.77\%	\\
\bottomrule
	\end{tabular}
\end{table*}

\begin{table*}[th]
	\caption{
		Results of state-of-the-art machine learning based methods and local search approaches (200/500/1000 iterations) on the Uchoa benchmark\cite{uchoa_bench}. For all instances sizes we use the NeuroLS model trained on instances of size 100. Percentages are the average gap to the best known upper bound. Best gap is marked in \textbf{bold}.
	}
	\label{appx:tab:results_cvrp}
	\centering
	\begin{tabular}{l|rrrrrrrr|r}
		\toprule
		\textbf{Model} & \multicolumn{8}{c|}{\textbf{Instance Group}} & \textbf{Avg} \\
		\hline
		& \multicolumn{2}{c}{\textit{n100}} & \multicolumn{2}{c}{\textit{n150}} & \multicolumn{2}{c}{\textit{n200}} & \multicolumn{2}{c|}{\textit{n250}} &   \\
		& cost & time & cost & time & cost & time & cost & time & \\
		\midrule
		\multicolumn{10}{c}{\textbf{constructive}} \\
		POMO\cite{kwon} (single) &	 23.49\%	&	 0.3	&	15.65\% 	&	0.4	&	25.86\% 	&	0.4 	&	22.93\% 	&	 0.4	&	 21.98\%	\\
		POMO\cite{kwon} &	 17.24\%	&	0.3 	&	 12.81\%	&	0.4	&	 20.52\%	&	0.4 	&	 15.14\%	&	 0.4	&	 16.43\%	\\
		POMO\cite{kwon} (aug) &	 6.32\%	&	\textbf{0.1} 	&	9.41\% 	&	\textbf{0.1}	&	13.47\% 	&	\textbf{0.1} 	&	 9.21\%	&	\textbf{0.1} 	&	9.60\% 	\\
		\midrule
		\multicolumn{10}{c}{\textbf{iterative (200)}} \\
\bottomrule
	\end{tabular}
\end{table*}

\clearpage
\subsection{Detailed Benchmark Results}\label{appx:benchmark}
In this last section we give the per instance results on the Taillard and Uchoa benchmark datasets for different methods and different number of iterations.

\begin{table*}[th]
	\caption{Detailed Results on the Taillard benchmark\cite{taillard_bench}.}
	\label{appx:tab:results_ta_11}
	\centering
	\begin{tabular}{ll|c|rrrrrrr}
		\textbf{Instance}	&		&	\textbf{BKS}	&	\multicolumn{7}{c}{\textbf{PDRs}}	\\
		 &		&		&	RND	&	FIFO	&	SPT	&	MWKR	&	MOPNR 	& FDD 	& $\frac{\text{FDD}}{\text{MWKR}}$\\
		 15x15 & \multicolumn{9}{c}{}		\\
		\midrule
		
		ta01	&	cost	&	1231	&	1655	&	1486	&	1462	&	1491	&	1438	&	1439	&	1417	\\
		&	time	&		&	0.04	&	0.03	&	0.03	&	0.03	&	0.03	&	0.03	&	0.03	\\
		ta02	&	cost	&	1244	&	1682	&	1486	&	1450	&	1449	&	1452	&	1447	&	1413	\\
		&	time	&		&	0.04	&	0.04	&	0.03	&	0.03	&	0.03	&	0.03	&	0.03	\\
		ta03	&	cost	&	1218	&	1596	&	1461	&	1526	&	1426	&	1418	&	1618	&	1423	\\
\bottomrule
	\end{tabular}
	
\end{table*}

\begin{table*}[th]
	\caption{Detailed Results on the Taillard benchmark\cite{taillard_bench}.}
	\label{appx:tab:results_ta_12}
	\centering
	\begin{tabular}{ll|c|rrrrrrr}
		\textbf{Instance}	&		&	\textbf{BKS}	&	\multicolumn{7}{c}{\textbf{PDRs}}	\\
		&		&		&	RND	&	FIFO	&	SPT	&	MWKR	&	MOPNR 	& FDD 	& $\frac{\text{FDD}}{\text{MWKR}}$\\
		20x20 & \multicolumn{9}{c}{}		\\
		\midrule
		ta21	&	cost	&	1642	&	2075	&	2089	&	2175	&	2044	&	1964	&	1962	&	1930	\\
		&	time	&		&	0.07	&	0.06	&	0.06	&	0.06	&	0.06	&	0.06	&	0.05	\\
		ta22	&	cost	&	1600	&	1985	&	2146	&	1893	&	1914	&	1905	&	1921	&	1928	\\
		&	time	&		&	0.07	&	0.06	&	0.06	&	0.05	&	0.05	&	0.06	&	0.06	\\
		ta23	&	cost	&	1557	&	2194	&	2010	&	1990	&	1954	&	1922	&	1945	&	1833	\\
\bottomrule
	\end{tabular}
	
\end{table*}

\begin{table*}[th]
	\caption{Detailed Results on the Taillard benchmark\cite{taillard_bench}.}
	\label{appx:tab:results_ta_13}
	\centering
	\begin{tabular}{ll|c|rrrrrrr}
		\textbf{Instance}	&		&	\textbf{BKS}	&	\multicolumn{7}{c}{\textbf{PDRs}}	\\
		&		&		&	RND	&	FIFO	&	SPT	&	MWKR	&	MOPNR 	& FDD 	& $\frac{\text{FDD}}{\text{MWKR}}$\\
		30x20 & \multicolumn{9}{c}{}		\\
		\midrule
		ta41	&	cost	&	2005	&	2863	&	2556	&	2463	&	2620	&	2538	&	2445	&	2537	\\
		&	time	&		&	0.11	&	0.08	&	0.09	&	0.09	&	0.09	&	0.09	&	0.09	\\
		ta42	&	cost	&	1937	&	2664	&	2648	&	2773	&	2424	&	2448	&	2567	&	2388	\\
		&	time	&		&	0.11	&	0.08	&	0.09	&	0.08	&	0.08	&	0.09	&	0.09	\\
		ta43	&	cost	&	1846	&	2468	&	2442	&	2412	&	2348	&	2425	&	2269	&	2354	\\
\bottomrule
	\end{tabular}
	
\end{table*}

\begin{table*}[th]
	\caption{Detailed Results on the Taillard benchmark\cite{taillard_bench}.}
	\label{appx:tab:results_ta_14}
	\centering
	\begin{tabular}{ll|c|rrrrrrr}
		\textbf{Instance}	&		&	\textbf{BKS}	&	\multicolumn{7}{c}{\textbf{PDRs}}	\\
		&		&		&	RND	&	FIFO	&	SPT	&	MWKR	&	MOPNR 	& FDD 	& $\frac{\text{FDD}}{\text{MWKR}}$\\
		50x20 & \multicolumn{9}{c}{}		\\
		\midrule
		ta61	&	cost	&	2868	&	3523	&	3690	&	3587	&	3343	&	3329	&	3303	&	3366	\\
		&	time	&		&	0.22	&	0.15	&	0.15	&	0.15	&	0.16	&	0.15	&	0.15	\\
		ta62	&	cost	&	2869	&	3810	&	3672	&	3487	&	3423	&	3417	&	3415	&	3437	\\
		&	time	&		&	0.2	&	0.15	&	0.17	&	0.15	&	0.16	&	0.15	&	0.16	\\
		ta63	&	cost	&	2755	&	3408	&	3320	&	3277	&	3233	&	3226	&	3247	&	3139	\\
\bottomrule
	\end{tabular}
	
\end{table*}

\clearpage
\newgeometry{top=30mm, bottom=25mm, left=45mm, right=45mm}
\begin{landscape}
\begin{table*}[th]
	\caption{Detailed Results on the Taillard benchmark\cite{taillard_bench}.}
	\label{appx:tab:results_ta_21}
	\centering
	\begin{tabular}{ll|rrr|rrr|rrr|rrr|rrr|rrr|rrr|rrr}
		\textbf{Inst.}	&		\multicolumn{24}{c}{\textbf{LS}}	\\
			& &	\multicolumn{3}{l}{SA} &	\multicolumn{3}{l}{SA$_{\text{re}}$} &	\multicolumn{3}{l}{ILS} &	\multicolumn{3}{l}{ILS+SA} &	\multicolumn{3}{l}{VNS} &	\multicolumn{3}{l}{NLS$_{\text{A}}$} &	\multicolumn{3}{l}{NLS$_{\text{AN}}$} &	\multicolumn{3}{l}{NLS$_{\text{ANP}}$} \\	
		15x15 &	&	50 &	100 &	200 & 	50 &	100 &	200 & 	50 &	100 &	200 & 	50 &	100 &	200 & 	50 &	100 &	200 & 	50 &	100 &	200 & 	50 &	100 &	200 & 	50 &	100 &	200 \\
		\midrule
		ta01	&	cost	&	1378	&	1378	&	1378	&	1378	&	1378	&	1378	&	1365	&	1352	&	1352	&	1365	&	1365	&	1365	&	1361	&	1361	&	1361	&	1346	&	1346	&	1320	&	1359	&	1339	&	1331	&	1365	&	1365	&	1350	\\
		&	time	&	0.39	&	0.78	&	1.57	&	0.39	&	0.81	&	1.66	&	0.43	&	0.87	&	1.88	&	0.42	&	0.78	&	1.5	&	1	&	2.18	&	4.74	&	0.88	&	1.62	&	2.96	&	1.03	&	1.89	&	4.01	&	1.07	&	1.94	&	3.75	\\
		ta02	&	cost	&	1387	&	1387	&	1386	&	1387	&	1387	&	1387	&	1376	&	1351	&	1332	&	1376	&	1376	&	1376	&	1361	&	1326	&	1326	&	1333	&	1311	&	1311	&	1369	&	1369	&	1326	&	1369	&	1369	&	1369	\\
		&	time	&	0.43	&	0.79	&	1.58	&	0.4	&	0.82	&	1.64	&	0.47	&	0.97	&	1.93	&	0.43	&	0.9	&	1.77	&	0.96	&	2.02	&	4.51	&	0.87	&	1.56	&	3.07	&	1.02	&	1.92	&	3.67	&	1.04	&	1.94	&	3.85	\\
		ta03	&	cost	&	1391	&	1391	&	1383	&	1391	&	1391	&	1391	&	1333	&	1332	&	1321	&	1353	&	1333	&	1333	&	1373	&	1330	&	1330	&	1334	&	1333	&	1318	&	1333	&	1333	&	1333	&	1353	&	1339	&	1339	\\
\bottomrule
	\end{tabular}
	
\end{table*}
\end{landscape}

\begin{landscape}
	\begin{table*}[th]
		\caption{Detailed Results on the Taillard benchmark\cite{taillard_bench}.}
		\label{appx:tab:results_ta_22}
		\centering
		\begin{tabular}{ll|rrr|rrr|rrr|rrr|rrr|rrr|rrr|rrr}
			\textbf{Inst.}	&		\multicolumn{24}{c}{\textbf{LS}}	\\
			& &	\multicolumn{3}{l}{SA} &	\multicolumn{3}{l}{SA$_{\text{re}}$} &	\multicolumn{3}{l}{ILS} &	\multicolumn{3}{l}{ILS+SA} &	\multicolumn{3}{l}{VNS} &	\multicolumn{3}{l}{NLS$_{\text{A}}$} &	\multicolumn{3}{l}{NLS$_{\text{AN}}$} &	\multicolumn{3}{l}{NLS$_{\text{ANP}}$} \\	
			20x15 &	&	50 &	100 &	200 & 	50 &	100 &	200 & 	50 &	100 &	200 & 	50 &	100 &	200 & 	50 &	100 &	200 & 	50 &	100 &	200 & 	50 &	100 &	200 & 	50 &	100 &	200 \\
			\midrule
			ta11	&	cost	&	1615	&	1615	&	1579	&	1615	&	1615	&	1615	&	1601	&	1579	&	1556	&	1611	&	1609	&	1579	&	1611	&	1579	&	1553	&	1575	&	1555	&	1555	&	1579	&	1568	&	1549	&	1608	&	1611	&	1611	\\
			&	time	&	0.55	&	1	&	2.12	&	0.51	&	1.09	&	2.23	&	0.66	&	1.28	&	2.68	&	0.6	&	1.17	&	2.47	&	1.42	&	2.86	&	5.97	&	1.38	&	2.08	&	4.44	&	1.16	&	2.26	&	5.08	&	1.31	&	3.21	&	6.79	\\
			ta12	&	cost	&	1642	&	1642	&	1611	&	1642	&	1642	&	1597	&	1571	&	1507	&	1485	&	1575	&	1571	&	1571	&	1594	&	1532	&	1509	&	1537	&	1512	&	1510	&	1542	&	1523	&	1472	&	1594	&	1594	&	1573	\\
			&	time	&	0.49	&	1.01	&	2.12	&	0.51	&	1.07	&	2.24	&	0.52	&	1.07	&	2.29	&	0.47	&	0.95	&	2.05	&	1.3	&	2.66	&	5.72	&	0.97	&	2.11	&	4.35	&	1.12	&	2.29	&	4.5	&	1.31	&	3.08	&	7.32	\\
			ta13	&	cost	&	1605	&	1556	&	1556	&	1605	&	1556	&	1556	&	1559	&	1548	&	1532	&	1564	&	1559	&	1558	&	1559	&	1559	&	1539	&	1541	&	1539	&	1496	&	1559	&	1541	&	1530	&	1556	&	1559	&	1559	\\
\bottomrule
		\end{tabular}
		
	\end{table*}
\end{landscape}

\begin{landscape}
	\begin{table*}[th]
		\caption{Detailed Results on the Taillard benchmark\cite{taillard_bench}.}
		\label{appx:tab:results_ta_23}
		\centering
		\begin{tabular}{ll|rrr|rrr|rrr|rrr|rrr|rrr|rrr|rrr}
			\textbf{Inst.}	&		\multicolumn{24}{c}{\textbf{LS}}	\\
			& &	\multicolumn{3}{l}{SA} &	\multicolumn{3}{l}{SA$_{\text{re}}$} &	\multicolumn{3}{l}{ILS} &	\multicolumn{3}{l}{ILS+SA} &	\multicolumn{3}{l}{VNS} &	\multicolumn{3}{l}{NLS$_{\text{A}}$} &	\multicolumn{3}{l}{NLS$_{\text{AN}}$} &	\multicolumn{3}{l}{NLS$_{\text{ANP}}$} \\	
			20x20 &	&	50 &	100 &	200 & 	50 &	100 &	200 & 	50 &	100 &	200 & 	50 &	100 &	200 & 	50 &	100 &	200 & 	50 &	100 &	200 & 	50 &	100 &	200 & 	50 &	100 &	200 \\
			\midrule
			ta21	&	cost	&	1930	&	1880	&	1865	&	1930	&	1930	&	1904	&	1835	&	1813	&	1795	&	1848	&	1835	&	1824	&	1871	&	1867	&	1858	&	1835	&	1819	&	1806	&	1835	&	1800	&	1803	&	1898	&	1863	&	1867	\\
			&	time	&	0.65	&	1.34	&	2.83	&	0.7	&	1.44	&	2.95	&	0.67	&	1.54	&	3.09	&	0.66	&	1.33	&	2.66	&	1.59	&	3.66	&	7.8	&	1.27	&	2.41	&	4.57	&	1.41	&	2.8	&	5.93	&	2.13	&	3.74	&	7.01	\\
			ta22	&	cost	&	1914	&	1871	&	1867	&	1914	&	1871	&	1871	&	1798	&	1796	&	1787	&	1812	&	1798	&	1796	&	1808	&	1808	&	1781	&	1789	&	1786	&	1772	&	1829	&	1774	&	1793	&	1911	&	1848	&	1821	\\
			&	time	&	0.66	&	1.33	&	2.74	&	0.65	&	1.32	&	2.83	&	0.65	&	1.46	&	3.11	&	0.59	&	1.3	&	2.82	&	1.46	&	3.71	&	7.53	&	1.19	&	2.38	&	4.75	&	1.36	&	2.76	&	6.16	&	2.28	&	3.7	&	7.01	\\
			ta23	&	cost	&	1811	&	1788	&	1786	&	1811	&	1791	&	1791	&	1745	&	1738	&	1722	&	1745	&	1745	&	1742	&	1796	&	1796	&	1757	&	1745	&	1745	&	1736	&	1765	&	1729	&	1736	&	1793	&	1804	&	1761	\\
\bottomrule
		\end{tabular}
		
	\end{table*}
\end{landscape}

\begin{landscape}
	\begin{table*}[th]
		\caption{Detailed Results on the Taillard benchmark\cite{taillard_bench}.}
		\label{appx:tab:results_ta_24}
		\centering
		\begin{tabular}{ll|rrr|rrr|rrr|rrr|rrr|rrr|rrr|rrr}
			\textbf{Inst.}	&		\multicolumn{24}{c}{\textbf{LS}}	\\
			& &	\multicolumn{3}{l}{SA} &	\multicolumn{3}{l}{SA$_{\text{re}}$} &	\multicolumn{3}{l}{ILS} &	\multicolumn{3}{l}{ILS+SA} &	\multicolumn{3}{l}{VNS} &	\multicolumn{3}{l}{NLS$_{\text{A}}$} &	\multicolumn{3}{l}{NLS$_{\text{AN}}$} &	\multicolumn{3}{l}{NLS$_{\text{ANP}}$} \\	
			30x15 &	&	50 &	100 &	200 & 	50 &	100 &	200 & 	50 &	100 &	200 & 	50 &	100 &	200 & 	50 &	100 &	200 & 	50 &	100 &	200 & 	50 &	100 &	200 & 	50 &	100 &	200 \\
			\midrule
			ta31	&	cost	&	2089	&	2081	&	2049	&	2089	&	2089	&	2089	&	2085	&	2077	&	2058	&	2089	&	2089	&	2077	&	2072	&	2072	&	2026	&	2072	&	2049	&	2001	&	2072	&	2067	&	2046	&	2081	&	2079	&	2048	\\
			&	time	&	0.81	&	1.56	&	3.12	&	0.76	&	1.62	&	3.53	&	0.91	&	1.83	&	3.65	&	0.84	&	1.57	&	3.35	&	1.93	&	4.2	&	8.78	&	1.43	&	2.72	&	5.26	&	1.64	&	3.45	&	7	&	1.77	&	4.24	&	6.77	\\
			ta32	&	cost	&	2148	&	2124	&	2065	&	2148	&	2124	&	2065	&	2125	&	2125	&	2125	&	2125	&	2125	&	2125	&	2125	&	2125	&	2125	&	2102	&	2065	&	2054	&	2125	&	2084	&	2058	&	2120	&	2087	&	2031	\\
			&	time	&	0.77	&	1.55	&	3.16	&	0.77	&	1.71	&	3.5	&	0.96	&	2.17	&	4.39	&	0.78	&	1.73	&	3.73	&	1.64	&	4.06	&	8.94	&	1.32	&	2.57	&	5.76	&	1.53	&	2.91	&	6.7	&	1.53	&	3.39	&	6.94	\\
			ta33	&	cost	&	2228	&	2200	&	2158	&	2228	&	2200	&	2172	&	2187	&	2154	&	2138	&	2255	&	2194	&	2154	&	2182	&	2134	&	2101	&	2169	&	2157	&	2133	&	2206	&	2130	&	2125	&	2169	&	2153	&	2099	\\
\bottomrule
		\end{tabular}
		
	\end{table*}
\end{landscape}

\begin{landscape}
	\begin{table*}[th]
		\caption{Detailed Results on the Taillard benchmark\cite{taillard_bench}.}
		\label{appx:tab:results_ta_25}
		\centering
		\begin{tabular}{ll|rrr|rrr|rrr|rrr|rrr|rrr|rrr|rrr}
			\textbf{Inst.}	&		\multicolumn{24}{c}{\textbf{LS}}	\\
			& &	\multicolumn{3}{l}{SA} &	\multicolumn{3}{l}{SA$_{\text{re}}$} &	\multicolumn{3}{l}{ILS} &	\multicolumn{3}{l}{ILS+SA} &	\multicolumn{3}{l}{VNS} &	\multicolumn{3}{l}{NLS$_{\text{A}}$} &	\multicolumn{3}{l}{NLS$_{\text{AN}}$} &	\multicolumn{3}{l}{NLS$_{\text{ANP}}$} \\	
			30x20 &	&	50 &	100 &	200 & 	50 &	100 &	200 & 	50 &	100 &	200 & 	50 &	100 &	200 & 	50 &	100 &	200 & 	50 &	100 &	200 & 	50 &	100 &	200 & 	50 &	100 &	200 \\
			\midrule
			ta41	&	cost	&	2490	&	2465	&	2429	&	2490	&	2465	&	2465	&	2427	&	2405	&	2404	&	2465	&	2427	&	2405	&	2465	&	2442	&	2442	&	2423	&	2414	&	2384	&	2434	&	2413	&	2372	&	2430	&	2408	&	2351	\\
			&	time	&	0.95	&	2.02	&	4.12	&	1.08	&	2.12	&	4.37	&	1.03	&	2.28	&	4.92	&	1.07	&	2.16	&	4.52	&	2.42	&	5.13	&	11.14	&	1.68	&	3.39	&	7.39	&	1.88	&	4.1	&	8.26	&	1.88	&	3.89	&	7.94	\\
			ta42	&	cost	&	2355	&	2331	&	2309	&	2374	&	2354	&	2307	&	2341	&	2330	&	2302	&	2341	&	2341	&	2333	&	2315	&	2277	&	2277	&	2291	&	2291	&	2254	&	2295	&	2295	&	2285	&	2291	&	2241	&	2233	\\
			&	time	&	0.98	&	2.03	&	4.16	&	1.29	&	2.15	&	4.3	&	1.25	&	2.61	&	5.16	&	1.08	&	2.26	&	4.7	&	2.38	&	5.13	&	11.29	&	1.63	&	3.39	&	7.29	&	1.93	&	4.14	&	8.44	&	1.96	&	3.95	&	8.61	\\
			ta43	&	cost	&	2354	&	2342	&	2308	&	2354	&	2296	&	2282	&	2297	&	2248	&	2247	&	2297	&	2297	&	2248	&	2268	&	2251	&	2216	&	2259	&	2228	&	2208	&	2272	&	2253	&	2201	&	2265	&	2232	&	2182	\\
\bottomrule
		\end{tabular}
		
	\end{table*}
\end{landscape}

\begin{landscape}
	\begin{table*}[th]
		\caption{Detailed Results on the Taillard benchmark\cite{taillard_bench}.}
		\label{appx:tab:results_ta_26}
		\centering
		\begin{tabular}{ll|rrr|rrr|rrr|rrr|rrr|rrr|rrr|rrr}
			\textbf{Inst.}	&		\multicolumn{24}{c}{\textbf{LS}}	\\
			& &	\multicolumn{3}{l}{SA} &	\multicolumn{3}{l}{SA$_{\text{re}}$} &	\multicolumn{3}{l}{ILS} &	\multicolumn{3}{l}{ILS+SA} &	\multicolumn{3}{l}{VNS} &	\multicolumn{3}{l}{NLS$_{\text{A}}$} &	\multicolumn{3}{l}{NLS$_{\text{AN}}$} &	\multicolumn{3}{l}{NLS$_{\text{ANP}}$} \\	
			50x15 &	&	50 &	100 &	200 & 	50 &	100 &	200 & 	50 &	100 &	200 & 	50 &	100 &	200 & 	50 &	100 &	200 & 	50 &	100 &	200 & 	50 &	100 &	200 & 	50 &	100 &	200 \\
			\midrule
			ta51	&	cost	&	3444	&	3381	&	3320	&	3444	&	3381	&	3320	&	3391	&	3355	&	3333	&	3391	&	3391	&	3355	&	3326	&	3287	&	3283	&	3391	&	3314	&	3273	&	3387	&	3300	&	3281	&	3373	&	3361	&	3281	\\
			&	time	&	1.34	&	2.65	&	5.51	&	1.37	&	2.71	&	5.37	&	1.46	&	2.9	&	6.31	&	1.33	&	2.81	&	5.66	&	1.71	&	4.38	&	12.36	&	2.25	&	4.22	&	8.48	&	2.23	&	4.31	&	9.56	&	2.29	&	4.83	&	9.99	\\
			ta52	&	cost	&	3254	&	3225	&	3210	&	3254	&	3225	&	3210	&	3228	&	3224	&	3198	&	3276	&	3240	&	3228	&	3276	&	3193	&	3160	&	3267	&	3231	&	3188	&	3240	&	3231	&	3161	&	3231	&	3231	&	3205	\\
			&	time	&	1.33	&	2.62	&	5.25	&	1.31	&	2.64	&	5.21	&	1.48	&	3.16	&	6.63	&	1.48	&	2.88	&	5.98	&	3.72	&	5.69	&	13.07	&	2.55	&	4.95	&	9.53	&	2.91	&	5.23	&	10.11	&	2.52	&	5.6	&	10.37	\\
			ta53	&	cost	&	2998	&	2987	&	2933	&	2998	&	2998	&	2998	&	3000	&	2994	&	2994	&	3014	&	3005	&	2994	&	2994	&	2968	&	2967	&	3010	&	2980	&	2974	&	2994	&	2990	&	2974	&	2983	&	2983	&	2974	\\
\bottomrule
		\end{tabular}
		
	\end{table*}
\end{landscape}

\begin{landscape}
	\begin{table*}[th]
		\caption{Detailed Results on the Taillard benchmark\cite{taillard_bench}.}
		\label{appx:tab:results_ta_27}
		\centering
		\begin{tabular}{ll|rrr|rrr|rrr|rrr|rrr|rrr|rrr|rrr}
			\textbf{Inst.}	&		\multicolumn{24}{c}{\textbf{LS}}	\\
			& &	\multicolumn{3}{l}{SA} &	\multicolumn{3}{l}{SA$_{\text{re}}$} &	\multicolumn{3}{l}{ILS} &	\multicolumn{3}{l}{ILS+SA} &	\multicolumn{3}{l}{VNS} &	\multicolumn{3}{l}{NLS$_{\text{A}}$} &	\multicolumn{3}{l}{NLS$_{\text{AN}}$} &	\multicolumn{3}{l}{NLS$_{\text{ANP}}$} \\	
			50x20 &	&	50 &	100 &	200 & 	50 &	100 &	200 & 	50 &	100 &	200 & 	50 &	100 &	200 & 	50 &	100 &	200 & 	50 &	100 &	200 & 	50 &	100 &	200 & 	50 &	100 &	200 \\
			\midrule
			ta61	&	cost	&	3252	&	3236	&	3205	&	3252	&	3236	&	3217	&	3215	&	3196	&	3175	&	3215	&	3196	&	3196	&	3204	&	3192	&	3192	&	3208	&	3167	&	3143	&	3208	&	3196	&	3181	&	3208	&	3188	&	3173	\\
			&	time	&	1.71	&	3.5	&	7	&	1.71	&	3.47	&	7.24	&	1.55	&	3.38	&	7.09	&	1.55	&	3.31	&	6.47	&	2.25	&	6.33	&	16.52	&	2.26	&	4.47	&	9.85	&	2.2	&	5.39	&	12.45	&	2.21	&	5.56	&	13.09	\\
			ta62	&	cost	&	3368	&	3337	&	3299	&	3368	&	3337	&	3299	&	3321	&	3321	&	3317	&	3321	&	3321	&	3321	&	3329	&	3329	&	3308	&	3329	&	3321	&	3296	&	3329	&	3295	&	3290	&	3322	&	3290	&	3269	\\
			&	time	&	1.66	&	3.44	&	7.02	&	1.67	&	3.42	&	7.19	&	1.72	&	3.59	&	7.99	&	1.7	&	3.28	&	6.5	&	3.31	&	8.29	&	18.25	&	2.79	&	5.74	&	12.01	&	2.96	&	5.69	&	12.79	&	3.09	&	5.89	&	13.02	\\
			ta63	&	cost	&	3095	&	3090	&	3065	&	3095	&	3090	&	3090	&	3076	&	3046	&	3030	&	3076	&	3076	&	3054	&	3076	&	3033	&	3025	&	3058	&	3049	&	3019	&	3058	&	3058	&	3011	&	3053	&	3028	&	3009	\\
\bottomrule
		\end{tabular}
		
	\end{table*}
\end{landscape}

\begin{landscape}
	\begin{table*}[th]
		\caption{Detailed Results on the Taillard benchmark\cite{taillard_bench}.}
		\label{appx:tab:results_ta_28}
		\centering
		\begin{tabular}{ll|rrr|rrr|rrr|rrr|rrr|rrr|rrr|rrr}
			\textbf{Inst.}	&		\multicolumn{24}{c}{\textbf{LS}}	\\
			& &	\multicolumn{3}{l}{SA} &	\multicolumn{3}{l}{SA$_{\text{re}}$} &	\multicolumn{3}{l}{ILS} &	\multicolumn{3}{l}{ILS+SA} &	\multicolumn{3}{l}{VNS} &	\multicolumn{3}{l}{NLS$_{\text{A}}$} &	\multicolumn{3}{l}{NLS$_{\text{AN}}$} &	\multicolumn{3}{l}{NLS$_{\text{ANP}}$} \\	
			 &	&	50 &	100 &	200 & 	50 &	100 &	200 & 	50 &	100 &	200 & 	50 &	100 &	200 & 	50 &	100 &	200 & 	50 &	100 &	200 & 	50 &	100 &	200 & 	50 &	100 &	200 \\
			\multicolumn{2}{l}{100x20} & \multicolumn{23}{c}{}\\
			\midrule
			ta71	&	cost	&	5855	&	5822	&	5813	&	5855	&	5855	&	5807	&	5855	&	5855	&	5855	&	5855	&	5855	&	5855	&	5848	&	5820	&	5810	&	5832	&	5817	&	5811	&	5855	&	5821	&	5811	&	5842	&	5810	&	5799	\\
			&	time	&	3.79	&	7.22	&	14.21	&	4.11	&	8.33	&	15.63	&	4.85	&	9.83	&	19.8	&	3.94	&	8.1	&	16.26	&	9.97	&	18.47	&	38.01	&	6.06	&	11.69	&	24.48	&	7.48	&	13.23	&	26.77	&	7.17	&	12.63	&	25.61	\\
			ta72	&	cost	&	5531	&	5528	&	5490	&	5531	&	5531	&	5531	&	5499	&	5474	&	5460	&	5499	&	5499	&	5480	&	5496	&	5496	&	5458	&	5480	&	5471	&	5446	&	5480	&	5465	&	5448	&	5480	&	5445	&	5431	\\
			&	time	&	3.61	&	7.19	&	14.75	&	3.58	&	7.99	&	16.08	&	3.8	&	8.16	&	17.58	&	3.66	&	7.74	&	15.61	&	5.56	&	15.53	&	31.26	&	4.27	&	10.56	&	22.76	&	4.5	&	11.51	&	26.33	&	4.69	&	10.54	&	24.12	\\
\bottomrule
		\end{tabular}
		
	\end{table*}
\end{landscape}

\restoregeometry
\clearpage
\begin{table*}[th]
	\caption{Detailed Results on the Uchoa benchmark\cite{uchoa_bench}.}
	\label{appx:tab:results_uchoa_11}
	\centering
	\begin{tabular}{ll|c|rrrrr}
		\textbf{Instance}	&		&	\textbf{BKS}	&	\multicolumn{3}{l}{\textbf{POMO}\cite{kwon}}		&	\textbf{VRPH}	&		\\
		n100 &		&		&	single	&	multi	&	aug=8	&	Sweep	&	Savings	\\
		\midrule
		X-n101-k25	&	cost	&	27591	&	29627	&	29025	&	28596	&	35807	&	29674	\\
		&	k	&		&	28	&	27	&	27	&	30	&	30	\\
		&	time	&		&	0.29	&	0.3	&	0.04	&	0	&	0	\\
		X-n106-k14	&	cost	&	26362	&	27521	&	27140	&	26849	&	34811	&	27442	\\
		&	k	&		&	14	&	14	&	14	&	14	&	14	\\
		&	time	&		&	0.28	&	0.29	&	0.04	&	0	&	0	\\
		X-n110-k13	&	cost	&	14971	&	15850	&	15316	&	15278	&	20435	&	15883	\\
		&	k	&		&	14	&	13	&	13	&	14	&	13	\\
		&	time	&		&	0.28	&	0.29	&	0.04	&	0	&	0	\\
		X-n115-k10	&	cost	&	12747	&	13523	&	13192	&	13192	&	18727	&	13585	\\
		&	k	&		&	10	&	10	&	10	&	11	&	11	\\
		&	time	&		&	0.28	&	0.29	&	0.04	&	0	&	0	\\
		X-n120-k6	&	cost	&	13332	&	14077	&	13846	&	13614	&	33166	&	14745	\\
		&	k	&		&	6	&	6	&	6	&	6	&	6	\\
		&	time	&		&	0.28	&	0.29	&	0.04	&	0	&	0	\\
		X-n125-k30	&	cost	&	55539	&	79994	&	74465	&	59504	&	71695	&	60561	\\
		&	k	&		&	41	&	38	&	32	&	36	&	34	\\
		&	time	&		&	0.31	&	0.32	&	0.04	&	0	&	0	\\
		X-n129-k18	&	cost	&	28940	&	29576	&	29220	&	29220	&	48717	&	30407	\\
		&	k	&		&	18	&	18	&	18	&	19	&	18	\\
		&	time	&		&	0.3	&	0.3	&	0.04	&	0	&	0	\\
		X-n134-k13	&	cost	&	10916	&	11803	&	11601	&	11377	&	19640	&	11415	\\
		&	k	&		&	14	&	14	&	14	&	14	&	14	\\
		&	time	&		&	0.3	&	0.3	&	0.04	&	0	&	0	\\
		X-n139-k10	&	cost	&	13590	&	14714	&	14147	&	13900	&	22174	&	14726	\\
		&	k	&		&	11	&	11	&	10	&	11	&	11	\\
		&	time	&		&	0.3	&	0.3	&	0.04	&	0	&	0	\\
		X-n143-k7	&	cost	&	15700	&	16195	&	16138	&	16138	&	41910	&	17664	\\
		&	k	&		&	7	&	7	&	7	&	7	&	7	\\
		&	time	&		&	0.3	&	0.31	&	0.04	&	0	&	0	\\
		X-n148-k46	&	cost	&	43448	&	72057	&	64407	&	52085	&	61758	&	48048	\\
		&	k	&		&	77	&	66	&	52	&	57	&	53	\\
		&	time	&		&	0.34	&	0.35	&	0.05	&	0	&	0	\\
		\bottomrule
	\end{tabular}
	
\end{table*}

\begin{table*}[th]
	\caption{Detailed Results on the Uchoa benchmark\cite{uchoa_bench}.}
	\label{appx:tab:results_uchoa_12}
	\centering
	\begin{tabular}{ll|c|rrrrr}
		\textbf{Instance}	&		&	\textbf{BKS}	&	\multicolumn{3}{l}{\textbf{POMO}\cite{kwon}}		&	\textbf{VRPH}	&		\\
		n150 &		&		&	single	&	multi	&	aug=8	&	Sweep	&	Savings	\\
		\midrule
		X-n153-k22	&	cost	&	21220	&	25709	&	24591	&	23833	&	31639	&	22649	\\
		&	k	&		&	27	&	26	&	26	&	29	&	26	\\
		&	time	&		&	0.67	&	0.39	&	0.05	&	0	&	0	\\
		X-n157-k13	&	cost	&	16876	&	17544	&	17331	&	17164	&	25353	&	19171	\\
		&	k	&		&	13	&	13	&	13	&	15	&	15	\\
		&	time	&		&	0.33	&	0.43	&	0.04	&	0	&	0	\\
		X-n162-k11	&	cost	&	14138	&	15142	&	14844	&	14813	&	24262	&	15708	\\
		&	k	&		&	11	&	11	&	11	&	11	&	11	\\
		&	time	&		&	0.32	&	0.37	&	0.04	&	0	&	0	\\
		X-n167-k10	&	cost	&	20557	&	22039	&	21564	&	21390	&	47099	&	22004	\\
		&	k	&		&	10	&	10	&	10	&	10	&	10	\\
		&	time	&		&	0.33	&	0.35	&	0.04	&	0	&	0	\\
		X-n172-k51	&	cost	&	45607	&	61768	&	60867	&	55635	&	61324	&	48634	\\
		&	k	&		&	69	&	66	&	56	&	63	&	57	\\
		&	time	&		&	0.37	&	0.39	&	0.05	&	0	&	0	\\
		X-n176-k26	&	cost	&	47812	&	55315	&	53233	&	52723	&	79892	&	52866	\\
		&	k	&		&	28	&	28	&	28	&	33	&	29	\\
		&	time	&		&	0.35	&	0.36	&	0.05	&	0	&	0	\\
		X-n181-k23	&	cost	&	25569	&	26255	&	26101	&	26101	&	33162	&	26561	\\
		&	k	&		&	23	&	23	&	23	&	23	&	23	\\
		&	time	&		&	0.35	&	0.38	&	0.05	&	0	&	0	\\
		X-n186-k15	&	cost	&	24145	&	25673	&	24683	&	24683	&	48964	&	25690	\\
		&	k	&		&	15	&	15	&	15	&	15	&	15	\\
		&	time	&		&	0.35	&	0.39	&	0.05	&	0	&	0	\\
		X-n190-k8	&	cost	&	16980	&	20053	&	19642	&	18550	&	40624	&	17816	\\
		&	k	&		&	8	&	8	&	8	&	8	&	8	\\
		&	time	&		&	0.55	&	0.38	&	0.05	&	0	&	0	\\
		X-n195-k51	&	cost	&	44225	&	51016	&	49771	&	48308	&	60623	&	45976	\\
		&	k	&		&	56	&	55	&	54	&	62	&	55	\\
		&	time	&		&	0.41	&	0.46	&	0.06	&	0	&	0	\\
		\bottomrule
	\end{tabular}
	
\end{table*}

\begin{table*}[th]
	\caption{Detailed Results on the Uchoa benchmark\cite{uchoa_bench}.}
	\label{appx:tab:results_uchoa_13}
	\centering
	\begin{tabular}{ll|c|rrrrr}
		\textbf{Instance}	&		&	\textbf{BKS}	&	\multicolumn{3}{l}{\textbf{POMO}\cite{kwon}}		&	\textbf{VRPH}	&		\\
		n200 &		&		&	single	&	multi	&	aug=8	&	Sweep	&	Savings	\\
		\midrule
		X-n200-k36	&	cost	&	58578	&	63316	&	61811	&	61512	&	72574	&	61911	\\
		&	k	&		&	37	&	37	&	37	&	39	&	38	\\
		&	time	&		&	0.39	&	0.38	&	0.06	&	0	&	0	\\
		X-n204-k19	&	cost	&	19565	&	21581	&	20853	&	20657	&	33405	&	21004	\\
		&	k	&		&	20	&	19	&	19	&	19	&	19	\\
		&	time	&		&	0.37	&	0.37	&	0.06	&	0	&	0	\\
		X-n209-k16	&	cost	&	30656	&	33346	&	32316	&	32145	&	66836	&	32793	\\
		&	k	&		&	16	&	16	&	16	&	17	&	16	\\
		&	time	&		&	0.36	&	0.38	&	0.05	&	0	&	0	\\
		X-n214-k11	&	cost	&	10856	&	17074	&	14923	&	14129	&	25760	&	11949	\\
		&	k	&		&	20	&	15	&	14	&	12	&	12	\\
		&	time	&		&	0.38	&	0.38	&	0.06	&	0	&	0	\\
		X-n219-k73	&	cost	&	117595	&	164917	&	157869	&	148715	&	199383	&	172876	\\
		&	k	&		&	104	&	101	&	87	&	109	&	109	\\
		&	time	&		&	0.44	&	0.44	&	0.07	&	0	&	0	\\
		X-n223-k34	&	cost	&	40437	&	42884	&	42162	&	42161	&	56590	&	42639	\\
		&	k	&		&	34	&	35	&	35	&	37	&	36	\\
		&	time	&		&	0.4	&	0.4	&	0.06	&	0	&	0	\\
		X-n228-k23	&	cost	&	25742	&	30048	&	28727	&	27596	&	44883	&	27296	\\
		&	k	&		&	24	&	24	&	23	&	29	&	25	\\
		&	time	&		&	0.38	&	0.4	&	0.06	&	0	&	0	\\
		X-n233-k16	&	cost	&	19230	&	27044	&	25061	&	20983	&	33749	&	20585	\\
		&	k	&		&	17	&	17	&	17	&	17	&	17	\\
		&	time	&		&	0.38	&	0.4	&	0.06	&	0	&	0	\\
		X-n237-k14	&	cost	&	27042	&	29262	&	29041	&	28835	&	71726	&	30571	\\
		&	k	&		&	14	&	14	&	14	&	14	&	14	\\
		&	time	&		&	0.37	&	0.4	&	0.07	&	0	&	0	\\
		X-n242-k48	&	cost	&	82751	&	85689	&	85463	&	85325	&	123830	&	89918	\\
		&	k	&		&	48	&	48	&	48	&	55	&	52	\\
		&	time	&		&	0.4	&	0.41	&	0.07	&	0	&	0.01	\\
		X-n247-k50	&	cost	&	37274	&	76015	&	67888	&	50926	&	56494	&	41935	\\
		&	k	&		&	94	&	81	&	58	&	65	&	59	\\
		&	time	&		&	0.43	&	0.47	&	0.08	&	0	&	0.01	\\
		\bottomrule
	\end{tabular}
	
\end{table*}

\begin{table*}[th]
	\caption{Detailed Results on the Uchoa benchmark\cite{uchoa_bench}.}
	\label{appx:tab:results_uchoa_14}
	\centering
	\begin{tabular}{ll|c|rrrrr}
		\textbf{Instance}	&		&	\textbf{BKS}	&	\multicolumn{3}{l}{\textbf{POMO}\cite{kwon}}		&	\textbf{VRPH}	&		\\
		n250 &		&		&	single	&	multi	&	aug=8	&	Sweep	&	Savings	\\
		\midrule
		X-n251-k28	&	cost	&	38684	&	41465	&	40784	&	40466	&	60078	&	40285	\\
		&	k	&		&	28	&	28	&	28	&	29	&	28	\\
		&	time	&		&	0.43	&	0.43	&	0.07	&	0	&	0.01	\\
		X-n256-k16	&	cost	&	18839	&	39858	&	29117	&	22825	&	28514	&	20518	\\
		&	k	&		&	35	&	23	&	19	&	17	&	17	\\
		&	time	&		&	0.42	&	0.41	&	0.08	&	0	&	0.01	\\
		X-n261-k13	&	cost	&	26558	&	30107	&	29300	&	28842	&	70523	&	28638	\\
		&	k	&		&	13	&	13	&	13	&	13	&	13	\\
		&	time	&		&	0.39	&	0.4	&	0.07	&	0	&	0.01	\\
		X-n266-k58	&	cost	&	75478	&	81464	&	80122	&	80122	&	115703	&	81046	\\
		&	k	&		&	60	&	60	&	60	&	67	&	64	\\
		&	time	&		&	0.41	&	0.44	&	0.08	&	0	&	0.01	\\
		X-n270-k35	&	cost	&	35291	&	39142	&	38535	&	38367	&	48710	&	37181	\\
		&	k	&		&	36	&	37	&	37	&	38	&	37	\\
		&	time	&		&	0.41	&	0.45	&	0.08	&	0	&	0.01	\\
		X-n275-k28	&	cost	&	21245	&	28811	&	25833	&	23360	&	35048	&	24177	\\
		&	k	&		&	32	&	29	&	28	&	31	&	31	\\
		&	time	&		&	0.41	&	0.42	&	0.08	&	0	&	0.01	\\
		X-n280-k17	&	cost	&	33503	&	36961	&	36503	&	35835	&	88461	&	35951	\\
		&	k	&		&	17	&	17	&	17	&	21	&	18	\\
		&	time	&		&	0.41	&	0.41	&	0.08	&	0	&	0.01	\\
		X-n284-k15	&	cost	&	20215	&	47812	&	38875	&	25404	&	44890	&	21924	\\
		&	k	&		&	35	&	24	&	16	&	15	&	15	\\
		&	time	&		&	0.44	&	0.45	&	0.09	&	0	&	0.01	\\
		X-n289-k60	&	cost	&	95151	&	104199	&	101303	&	99850	&	146744	&	99808	\\
		&	k	&		&	66	&	64	&	63	&	70	&	65	\\
		&	time	&		&	0.44	&	0.54	&	0.09	&	0	&	0.01	\\
		X-n294-k50	&	cost	&	47161	&	58954	&	55642	&	55011	&	67614	&	49307	\\
		&	k	&		&	54	&	53	&	53	&	56	&	53	\\
		&	time	&		&	0.43	&	0.47	&	0.1	&	0	&	0.01	\\
		X-n298-k31	&	cost	&	34231	&	39956	&	37929	&	37363	&	57847	&	35996	\\
		&	k	&		&	32	&	32	&	32	&	33	&	32	\\
		&	time	&		&	0.43	&	0.44	&	0.09	&	0	&	0.01	\\
		\bottomrule
	\end{tabular}
	
\end{table*}

\newgeometry{top=30mm, bottom=25mm, left=40mm, right=40mm}
\clearpage
\begin{landscape}	
\begin{table*}[th]
		\caption{Detailed Results on the Uchoa benchmark\cite{uchoa_bench} (200 iterations).}
		\label{appx:tab:results_uchoa_21}
		\centering
		\begin{tabular}{ll|c|rr|rrrrr|rr|rrr}
			\textbf{Instance}	&		&	\textbf{BKS}	&	\multicolumn{2}{l}{\textbf{ORTools}\cite{perron}}		&	\multicolumn{5}{l}{\textbf{Meta-VRPH}}		&	\multicolumn{2}{l}{\textbf{DACT}\cite{ma}}	&	\multicolumn{3}{l}{\textbf{NeuroLS}}	\\
			n100 &		&		&	GLS	&	TS	&	SA	&	SA$_{\text{re}}$	&	ILS	&	ILS+SA	&	VNS	&	aug=1	&	aug=8	&	NLS$_{\text{A}}$	&	NLS$_{\text{AN}}$	&	NLS$_{\text{ANP}}$	\\
			\midrule
			
			X-n101-k25	&	cost	&	27591	&	28906	&	29804	&	29181	&	29243	&	28845	&	29243	&	29674	&	30927	&	29562	&	28506	&	28444	&	28444	\\
			&	k	&		&	26	&	26	&	30	&	30	&	27	&	30	&	30	&	27	&	28	&	27	&	27	&	27	\\
			&	time	&		&	2.09	&	5.73	&	0.59	&	0.47	&	0.5	&	0.43	&	0.27	&	12.37	&	13.36	&	1.07	&	1.23	&	1.25	\\
			X-n106-k14	&	cost	&	26362	&	27182	&	27182	&	27099	&	27099	&	27442	&	27099	&	27442	&	28323	&	28085	&	26928	&	27057	&	27057	\\
			&	k	&		&	14	&	14	&	14	&	14	&	14	&	14	&	14	&	14	&	14	&	14	&	14	&	14	\\
			&	time	&		&	2.53	&	5.52	&	0.45	&	0.48	&	0.54	&	0.44	&	0.27	&	12.89	&	14.01	&	1.17	&	1.31	&	1.36	\\
\bottomrule
		\end{tabular}
	
\end{table*}
\end{landscape}

\begin{landscape}	
	\begin{table*}[th]
		\caption{Detailed Results on the Uchoa benchmark\cite{uchoa_bench} (200 iterations).}
		\label{appx:tab:results_uchoa_22}
		\centering
		\begin{tabular}{ll|c|rr|rrrrr|rr|rrr}
			\textbf{Instance}	&		&	\textbf{BKS}	&	\multicolumn{2}{l}{\textbf{ORTools}\cite{perron}}		&	\multicolumn{5}{l}{\textbf{Meta-VRPH}}		&	\multicolumn{2}{l}{\textbf{DACT}\cite{ma}}	&	\multicolumn{3}{l}{\textbf{NeuroLS}}	\\
			n100 &		&		&	GLS	&	TS	&	SA	&	SA$_{\text{re}}$	&	ILS	&	ILS+SA	&	VNS	&	aug=1	&	aug=8	&	NLS$_{\text{A}}$	&	NLS$_{\text{AN}}$	&	NLS$_{\text{ANP}}$	\\
			\midrule
			
			X-n153-k22	&	cost	&	21220	&	24812	&	24812	&	22597	&	22597	&	22649	&	22597	&	22649	&	26294	&	25312	&	22587	&	22548	&	22548	\\
			&	k	&		&	25	&	25	&	26	&	26	&	26	&	26	&	26	&	26	&	25	&	26	&	26	&	26	\\
			&	time	&		&	0.74	&	0.76	&	1.12	&	1.01	&	1.27	&	0.93	&	0.49	&	19.78	&	20.39	&	2.09	&	2.24	&	2.21	\\
			X-n157-k13	&	cost	&	16876	&	17257	&	17286	&	18963	&	18963	&	19050	&	19171	&	19171	&	18556	&	18352	&	18782	&	18810	&	18810	\\
			&	k	&		&	13	&	13	&	15	&	15	&	15	&	15	&	15	&	13	&	13	&	15	&	15	&	15	\\
			&	time	&		&	13.77	&	10.37	&	0.96	&	1.01	&	1.2	&	0.86	&	0.49	&	19.44	&	20.86	&	1.98	&	2.16	&	2.08	\\
\bottomrule			
		\end{tabular}
		
	\end{table*}
\end{landscape}

\begin{landscape}	
	\begin{table*}[th]
		\caption{Detailed Results on the Uchoa benchmark\cite{uchoa_bench} (200 iterations).}
		\label{appx:tab:results_uchoa_23}
		\centering
		\begin{tabular}{ll|c|rr|rrrrr|rr|rrr}
			\textbf{Instance}	&		&	\textbf{BKS}	&	\multicolumn{2}{l}{\textbf{ORTools}\cite{perron}}		&	\multicolumn{5}{l}{\textbf{Meta-VRPH}}		&	\multicolumn{2}{l}{\textbf{DACT}\cite{ma}}	&	\multicolumn{3}{l}{\textbf{NeuroLS}}	\\
			n100 &		&		&	GLS	&	TS	&	SA	&	SA$_{\text{re}}$	&	ILS	&	ILS+SA	&	VNS	&	aug=1	&	aug=8	&	NLS$_{\text{A}}$	&	NLS$_{\text{AN}}$	&	NLS$_{\text{ANP}}$	\\
			\midrule
			
			X-n200-k36	&	cost	&	58578	&	61080	&	61080	&	60902	&	60913	&	61051	&	60920	&	61911	&	64205	&	64746	&	60373	&	60337	&	60337	\\
			&	k	&		&	37	&	37	&	38	&	38	&	38	&	38	&	38	&	37	&	38	&	37	&	37	&	37	\\
			&	time	&		&	4.99	&	9.32	&	1.57	&	1.49	&	1.65	&	1.35	&	0.66	&	24.31	&	26.05	&	2.76	&	3.03	&	3.12	\\
			X-n204-k19	&	cost	&	19565	&	21828	&	21922	&	20906	&	20970	&	21004	&	21004	&	21004	&	24188	&	23259	&	20925	&	20587	&	20587	\\
			&	k	&		&	19	&	19	&	19	&	19	&	19	&	19	&	19	&	20	&	19	&	19	&	19	&	19	\\
			&	time	&		&	25.3	&	12.72	&	1.71	&	1.58	&	2.02	&	1.55	&	0.7	&	25.05	&	28.41	&	2.88	&	3.26	&	3.41	\\
\bottomrule			
		\end{tabular}
		
	\end{table*}
\end{landscape}

\begin{landscape}	
	\begin{table*}[th]
		\caption{Detailed Results on the Uchoa benchmark\cite{uchoa_bench} (200 iterations).}
		\label{appx:tab:results_uchoa_24}
		\centering
		\begin{tabular}{ll|c|rr|rrrrr|rr|rrr}
			\textbf{Instance}	&		&	\textbf{BKS}	&	\multicolumn{2}{l}{\textbf{ORTools}\cite{perron}}		&	\multicolumn{5}{l}{\textbf{Meta-VRPH}}		&	\multicolumn{2}{l}{\textbf{DACT}\cite{ma}}	&	\multicolumn{3}{l}{\textbf{NeuroLS}}	\\
			n100 &		&		&	GLS	&	TS	&	SA	&	SA$_{\text{re}}$	&	ILS	&	ILS+SA	&	VNS	&	aug=1	&	aug=8	&	NLS$_{\text{A}}$	&	NLS$_{\text{AN}}$	&	NLS$_{\text{ANP}}$	\\
			\midrule
			
			X-n251-k28	&	cost	&	38684	&	40417	&	40424	&	40279	&	40285	&	40285	&	40285	&	40285	&	42297	&	42545	&	40285	&	40285	&	40285	\\
			&	k	&		&	28	&	28	&	28	&	28	&	28	&	28	&	28	&	28	&	28	&	28	&	28	&	28	\\
			&	time	&		&	14.7	&	32.3	&	2.63	&	2.23	&	2.82	&	2.09	&	0.95	&	30.17	&	34.44	&	4.04	&	4.73	&	4.5	\\
			X-n256-k16	&	cost	&	18839	&	21363	&	21361	&	20518	&	20518	&	20518	&	20518	&	20518	&	23949	&	22469	&	20390	&	20185	&	20185	\\
			&	k	&		&	17	&	17	&	17	&	17	&	17	&	17	&	17	&	17	&	17	&	17	&	17	&	17	\\
			&	time	&		&	5.47	&	5.27	&	3.11	&	2.66	&	3.94	&	2.49	&	1.24	&	31.66	&	35.04	&	4.98	&	5.31	&	5.54	\\
\bottomrule			
		\end{tabular}
		
	\end{table*}
\end{landscape}

\begin{landscape}	
	\begin{table*}[th]
		\caption{Detailed Results on the Uchoa benchmark\cite{uchoa_bench} (500 iterations).}
		\label{appx:tab:results_uchoa_31}
		\centering
		\begin{tabular}{ll|c|rr|rrrrr|rr|rrr}
			\textbf{Instance}	&		&	\textbf{BKS}	&	\multicolumn{2}{l}{\textbf{ORTools}\cite{perron}}		&	\multicolumn{5}{l}{\textbf{Meta-VRPH}}		&	\multicolumn{2}{l}{\textbf{DACT}\cite{ma}}	&	\multicolumn{3}{l}{\textbf{NeuroLS}}	\\
			n100 &		&		&	GLS	&	TS	&	SA	&	SA$_{\text{re}}$	&	ILS	&	ILS+SA	&	VNS	&	aug=1	&	aug=8	&	NLS$_{\text{A}}$	&	NLS$_{\text{AN}}$	&	NLS$_{\text{ANP}}$	\\
			\midrule
			
			X-n101-k25	&	cost	&	27591	&	28725	&	29430	&	29181	&	29211	&	28845	&	29243	&	29674	&	29561	&	29554	&	28458	&	28309	&	28309	\\
			&	k	&		&	26	&	26	&	30	&	30	&	27	&	30	&	30	&	27	&	27	&	27	&	27	&	27	\\
			&	time	&		&	18.79	&	134.68	&	1.08	&	1.06	&	1.31	&	1.01	&	0.71	&	31.32	&	34.52	&	2.32	&	2.75	&	2.74	\\
			X-n106-k14	&	cost	&	26362	&	27182	&	27182	&	27099	&	27099	&	27442	&	27099	&	27442	&	28052	&	28085	&	26875	&	26747	&	26747	\\
			&	k	&		&	14	&	14	&	14	&	14	&	14	&	14	&	14	&	14	&	14	&	14	&	14	&	14	\\
			&	time	&		&	20.28	&	38.39	&	1.15	&	1.19	&	1.34	&	1.14	&	0.77	&	32.58	&	36.42	&	2.62	&	2.96	&	2.79	\\
\bottomrule			
		\end{tabular}
		
	\end{table*}
\end{landscape}

\begin{landscape}	
	\begin{table*}[th]
		\caption{Detailed Results on the Uchoa benchmark\cite{uchoa_bench} (500 iterations).}
		\label{appx:tab:results_uchoa_32}
		\centering
		\begin{tabular}{ll|c|rr|rrrrr|rr|rrr}
			\textbf{Instance}	&		&	\textbf{BKS}	&	\multicolumn{2}{l}{\textbf{ORTools}\cite{perron}}		&	\multicolumn{5}{l}{\textbf{Meta-VRPH}}		&	\multicolumn{2}{l}{\textbf{DACT}\cite{ma}}	&	\multicolumn{3}{l}{\textbf{NeuroLS}}	\\
			n100 &		&		&	GLS	&	TS	&	SA	&	SA$_{\text{re}}$	&	ILS	&	ILS+SA	&	VNS	&	aug=1	&	aug=8	&	NLS$_{\text{A}}$	&	NLS$_{\text{AN}}$	&	NLS$_{\text{ANP}}$	\\
			\midrule
			
			X-n153-k22	&	cost	&	21220	&	22507	&	22670	&	22597	&	22597	&	22649	&	22597	&	22649	&	25709	&	24538	&	22587	&	22495	&	22495	\\
			&	k	&		&	25	&	25	&	26	&	26	&	26	&	26	&	26	&	25	&	24	&	26	&	25	&	25	\\
			&	time	&		&	19.98	&	327.65	&	2.55	&	2.52	&	3.16	&	2.3	&	1.25	&	46.37	&	51.29	&	4.78	&	5.23	&	5.5	\\
			X-n157-k13	&	cost	&	16876	&	17238	&	17286	&	18963	&	18963	&	19050	&	19171	&	19171	&	18556	&	18352	&	18782	&	18696	&	18696	\\
			&	k	&		&	13	&	13	&	15	&	15	&	15	&	15	&	15	&	13	&	13	&	15	&	15	&	15	\\
			&	time	&		&	47.88	&	564.82	&	2.31	&	2.32	&	2.93	&	2.11	&	1.19	&	48.52	&	52.49	&	4.51	&	5.04	&	5.55	\\
\bottomrule			
		\end{tabular}
		
	\end{table*}
\end{landscape}

\begin{landscape}	
	\begin{table*}[th]
		\caption{Detailed Results on the Uchoa benchmark\cite{uchoa_bench} (500 iterations).}
		\label{appx:tab:results_uchoa_33}
		\centering
		\begin{tabular}{ll|c|rr|rrrrr|rr|rrr}
			\textbf{Instance}	&		&	\textbf{BKS}	&	\multicolumn{2}{l}{\textbf{ORTools}\cite{perron}}		&	\multicolumn{5}{l}{\textbf{Meta-VRPH}}		&	\multicolumn{2}{l}{\textbf{DACT}\cite{ma}}	&	\multicolumn{3}{l}{\textbf{NeuroLS}}	\\
			n100 &		&		&	GLS	&	TS	&	SA	&	SA$_{\text{re}}$	&	ILS	&	ILS+SA	&	VNS	&	aug=1	&	aug=8	&	NLS$_{\text{A}}$	&	NLS$_{\text{AN}}$	&	NLS$_{\text{ANP}}$	\\
			\midrule
			
			X-n200-k36	&	cost	&	58578	&	60475	&	60865	&	60902	&	60913	&	61051	&	60920	&	61911	&	64112	&	64640	&	60316	&	60315	&	60315	\\
			&	k	&		&	37	&	37	&	38	&	38	&	38	&	38	&	38	&	37	&	37	&	37	&	37	&	37	\\
			&	time	&		&	62.28	&	1629.3	&	3.45	&	3.38	&	4.07	&	3.23	&	1.65	&	60.76	&	67.26	&	6.5	&	7.43	&	7.26	\\
			X-n204-k19	&	cost	&	19565	&	20897	&	21330	&	20906	&	20970	&	21004	&	21004	&	21004	&	24187	&	23045	&	20481	&	20587	&	20587	\\
			&	k	&		&	19	&	19	&	19	&	19	&	19	&	19	&	19	&	19	&	19	&	19	&	19	&	19	\\
			&	time	&		&	129.9	&	1503.12	&	3.84	&	3.74	&	4.83	&	3.61	&	1.77	&	61.87	&	67.71	&	6.84	&	7.8	&	8.12	\\
\bottomrule			
		\end{tabular}
		
	\end{table*}
\end{landscape}

\begin{landscape}	
	\begin{table*}[th]
		\caption{Detailed Results on the Uchoa benchmark\cite{uchoa_bench} (500 iterations).}
		\label{appx:tab:results_uchoa_34}
		\centering
		\begin{tabular}{ll|c|rr|rrrrr|rr|rrr}
			\textbf{Instance}	&		&	\textbf{BKS}	&	\multicolumn{2}{l}{\textbf{ORTools}\cite{perron}}		&	\multicolumn{5}{l}{\textbf{Meta-VRPH}}		&	\multicolumn{2}{l}{\textbf{DACT}\cite{ma}}	&	\multicolumn{3}{l}{\textbf{NeuroLS}}	\\
			n100 &		&		&	GLS	&	TS	&	SA	&	SA$_{\text{re}}$	&	ILS	&	ILS+SA	&	VNS	&	aug=1	&	aug=8	&	NLS$_{\text{A}}$	&	NLS$_{\text{AN}}$	&	NLS$_{\text{ANP}}$	\\
			\midrule
			
			X-n251-k28	&	cost	&	38684	&	40322	&	40215	&	40279	&	40285	&	40285	&	40285	&	40285	&	42297	&	42545	&	40071	&	40233	&	40233	\\
			&	k	&		&	28	&	28	&	28	&	28	&	28	&	28	&	28	&	28	&	28	&	28	&	28	&	28	\\
			&	time	&		&	107.2	&	1223.66	&	5.39	&	5.25	&	7.02	&	5.1	&	2.56	&	78.34	&	85.31	&	9.84	&	11.13	&	11.63	\\
			X-n256-k16	&	cost	&	18839	&	19856	&	20274	&	20518	&	20518	&	20518	&	20518	&	20518	&	23910	&	22469	&	20084	&	20021	&	20021	\\
			&	k	&		&	17	&	17	&	17	&	17	&	17	&	17	&	17	&	17	&	17	&	17	&	17	&	17	\\
			&	time	&		&	136.78	&	54.78	&	6.84	&	6.33	&	8.53	&	6.18	&	3.33	&	79.97	&	89.58	&	11.68	&	13.22	&	13.29	\\
\bottomrule			
		\end{tabular}
		
	\end{table*}
\end{landscape}

\begin{landscape}	
	\begin{table*}[th]
		\caption{Detailed Results on the Uchoa benchmark\cite{uchoa_bench} (1000 iterations).}
		\label{appx:tab:results_uchoa_41}
		\centering
		\begin{tabular}{ll|c|rr|rrrrr|rr|rrr}
			\textbf{Instance}	&		&	\textbf{BKS}	&	\multicolumn{2}{l}{\textbf{ORTools}\cite{perron}}		&	\multicolumn{5}{l}{\textbf{Meta-VRPH}}		&	\multicolumn{2}{l}{\textbf{DACT}\cite{ma}}	&	\multicolumn{3}{l}{\textbf{NeuroLS}}	\\
			n100 &		&		&	GLS	&	TS	&	SA	&	SA$_{\text{re}}$	&	ILS	&	ILS+SA	&	VNS	&	aug=1	&	aug=8	&	NLS$_{\text{A}}$	&	NLS$_{\text{AN}}$	&	NLS$_{\text{ANP}}$	\\
			\midrule
			
			X-n101-k25	&	cost	&	27591	&	28725	&	28614	&	29181	&	29211	&	28845	&	29243	&	29674	&	29322	&	29315	&	28414	&	28261	&	28261	\\
			&	k	&		&	26	&	26	&	30	&	30	&	27	&	30	&	30	&	27	&	27	&	27	&	26	&	26	\\
			&	time	&		&	52.8	&	231.94	&	2.03	&	2	&	2.39	&	2.04	&	1.35	&	67.91	&	67.28	&	4.75	&	5.88	&	5.31	\\
			X-n106-k14	&	cost	&	26362	&	27182	&	26789	&	27099	&	27099	&	27190	&	27099	&	27442	&	27885	&	28069	&	26656	&	26747	&	26747	\\
			&	k	&		&	14	&	14	&	14	&	14	&	14	&	14	&	14	&	14	&	14	&	14	&	14	&	14	\\
			&	time	&		&	45.25	&	165.72	&	2.21	&	2.37	&	2.67	&	2.31	&	1.44	&	67.19	&	70.63	&	5.37	&	6.68	&	5.7	\\
\bottomrule			
		\end{tabular}
		
	\end{table*}
\end{landscape}

\begin{landscape}	
	\begin{table*}[th]
		\caption{Detailed Results on the Uchoa benchmark\cite{uchoa_bench} (1000 iterations).}
		\label{appx:tab:results_uchoa_42}
		\centering
		\begin{tabular}{ll|c|rr|rrrrr|rr|rrr}
			\textbf{Instance}	&		&	\textbf{BKS}	&	\multicolumn{2}{l}{\textbf{ORTools}\cite{perron}}		&	\multicolumn{5}{l}{\textbf{Meta-VRPH}}		&	\multicolumn{2}{l}{\textbf{DACT}\cite{ma}}	&	\multicolumn{3}{l}{\textbf{NeuroLS}}	\\
			n100 &		&		&	GLS	&	TS	&	SA	&	SA$_{\text{re}}$	&	ILS	&	ILS+SA	&	VNS	&	aug=1	&	aug=8	&	NLS$_{\text{A}}$	&	NLS$_{\text{AN}}$	&	NLS$_{\text{ANP}}$	\\
			\midrule
			
			X-n153-k22	&	cost	&	21220	&	22441	&	22876	&	22597	&	22597	&	22328	&	22597	&	22649	&	24792	&	24260	&	22486	&	22495	&	22495	\\
			&	k	&		&	25	&	24	&	26	&	26	&	25	&	26	&	26	&	24	&	24	&	25	&	25	&	25	\\
			&	time	&		&	95.8	&	851.36	&	4.7	&	4.55	&	6.15	&	4.51	&	2.44	&	99.17	&	100.44	&	9.18	&	12.26	&	10.27	\\
			X-n157-k13	&	cost	&	16876	&	17209	&	17298	&	18963	&	18963	&	19050	&	19171	&	19171	&	18556	&	18352	&	18782	&	18696	&	18696	\\
			&	k	&		&	13	&	13	&	15	&	15	&	15	&	15	&	15	&	13	&	13	&	15	&	15	&	15	\\
			&	time	&		&	119.01	&	957.66	&	4.56	&	4.73	&	5.69	&	4.31	&	2.37	&	99.23	&	103.26	&	8.86	&	11.51	&	9.95	\\
\bottomrule			
		\end{tabular}
		
	\end{table*}
\end{landscape}

\begin{landscape}	
	\begin{table*}[th]
		\caption{Detailed Results on the Uchoa benchmark\cite{uchoa_bench} (1000 iterations).}
		\label{appx:tab:results_uchoa_43}
		\centering
		\begin{tabular}{ll|c|rr|rrrrr|rr|rrr}
			\textbf{Instance}	&		&	\textbf{BKS}	&	\multicolumn{2}{l}{\textbf{ORTools}\cite{perron}}		&	\multicolumn{5}{l}{\textbf{Meta-VRPH}}		&	\multicolumn{2}{l}{\textbf{DACT}\cite{ma}}	&	\multicolumn{3}{l}{\textbf{NeuroLS}}	\\
			n100 &		&		&	GLS	&	TS	&	SA	&	SA$_{\text{re}}$	&	ILS	&	ILS+SA	&	VNS	&	aug=1	&	aug=8	&	NLS$_{\text{A}}$	&	NLS$_{\text{AN}}$	&	NLS$_{\text{ANP}}$	\\
			\midrule
			
			X-n200-k36	&	cost	&	58578	&	60473	&	60810	&	60902	&	60913	&	61051	&	60920	&	61911	&	64112	&	64677	&	60290	&	60315	&	60315	\\
			&	k	&		&	37	&	37	&	38	&	38	&	38	&	38	&	38	&	37	&	37	&	37	&	37	&	37	\\
			&	time	&		&	156.86	&	1853.73	&	6.43	&	6.18	&	8.23	&	6.2	&	3.39	&	125.59	&	130.62	&	12.69	&	16.04	&	13.9	\\
			X-n204-k19	&	cost	&	19565	&	20840	&	21045	&	20906	&	20970	&	21004	&	21004	&	21004	&	23895	&	22868	&	20481	&	20587	&	20587	\\
			&	k	&		&	19	&	19	&	19	&	19	&	19	&	19	&	19	&	19	&	19	&	19	&	19	&	19	\\
			&	time	&		&	297.17	&	2850.43	&	7.32	&	7.21	&	9.61	&	7.17	&	3.69	&	129.61	&	135.19	&	13.58	&	18.25	&	15.7	\\
\bottomrule			
		\end{tabular}
		
	\end{table*}
\end{landscape}

\begin{landscape}	
	\begin{table*}[th]
		\caption{Detailed Results on the Uchoa benchmark\cite{uchoa_bench} (1000 iterations).}
		\label{appx:tab:results_uchoa_44}
		\centering
		\begin{tabular}{ll|c|rr|rrrrr|rr|rrr}
			\textbf{Instance}	&		&	\textbf{BKS}	&	\multicolumn{2}{l}{\textbf{ORTools}\cite{perron}}		&	\multicolumn{5}{l}{\textbf{Meta-VRPH}}		&	\multicolumn{2}{l}{\textbf{DACT}\cite{ma}}	&	\multicolumn{3}{l}{\textbf{NeuroLS}}	\\
			n100 &		&		&	GLS	&	TS	&	SA	&	SA$_{\text{re}}$	&	ILS	&	ILS+SA	&	VNS	&	aug=1	&	aug=8	&	NLS$_{\text{A}}$	&	NLS$_{\text{AN}}$	&	NLS$_{\text{ANP}}$	\\
			\midrule
			
			X-n251-k28	&	cost	&	38684	&	40320	&	40215	&	40279	&	40285	&	40285	&	40285	&	40285	&	42297	&	42540	&	40015	&	40062	&	40062	\\
			&	k	&		&	28	&	28	&	28	&	28	&	28	&	28	&	28	&	28	&	28	&	28	&	28	&	28	\\
			&	time	&		&	269.72	&	2767.31	&	10.44	&	10.37	&	13.78	&	9.92	&	5.15	&	155.26	&	166.65	&	19.54	&	25.48	&	22.31	\\
			X-n256-k16	&	cost	&	18839	&	19464	&	20274	&	20518	&	20518	&	20518	&	20518	&	20518	&	23292	&	22172	&	19845	&	20021	&	20021	\\
			&	k	&		&	17	&	17	&	17	&	17	&	17	&	17	&	17	&	17	&	17	&	17	&	17	&	17	\\
			&	time	&		&	422.55	&	1168.37	&	12.65	&	12.26	&	17.36	&	12.32	&	6.45	&	154.69	&	172.04	&	22.68	&	29.79	&	26.32	\\
\bottomrule			
		\end{tabular}
		
	\end{table*}
\end{landscape}

\restoregeometry
\clearpage


\end{document}